\documentclass[a4paper,fleqn]{cas-dc}

\usepackage[numbers]{natbib}
\usepackage{rotating}
\usepackage{pdflscape}
\usepackage{subfig}
\usepackage{graphicx}
\usepackage{hyperref}
\def\tsc#1{\csdef{#1}{\textsc{\lowercase{#1}}\xspace}}
\tsc{WGM}
\tsc{QE}
\tsc{EP}
\tsc{PMS}
\tsc{BEC}
\tsc{DE}

\begin{document}
\let\WriteBookmarks\relax
\def\floatpagepagefraction{1}
\def\textpagefraction{.001}

\shorttitle{Multi-Agent Reinforcement Learning in Intelligent Transportation Systems: A Comprehensive Survey}

\shortauthors{Donatus et~al.}

\title [mode = title]{Multi-Agent Reinforcement Learning in Intelligent Transportation Systems: A Comprehensive Survey}                      



%
\author[1]{Rexcharles Enyinna Donatus}[
        ]
\ead{rdonatus@afit.edu.ng}
\credit{Data curation}
\author[1]{Kumater Ter}[%
   ]
\ead{kumater.ter@afit.edu.ng}
\credit{Data curation}
\author[1]{Daniel Udekwe}[
                        orcid=0000-0003-1771-5320
                        ]

\cormark[1]


\ead{daudekwe@afit.edu.ng}


\credit{Conceptualization of this study, Methodology, Software, Writing - Original draft preparation}





\affiliation[1]{organization={Department of Aerospace of Engineering, Faculty of Air Engineering, Air Force Institute of Technology},
    city={Kaduna},
    country={Nigeria}}

\cortext[cor1]{Corresponding author}



\begin{abstract}
The growing complexity of urban mobility and the demand for efficient, sustainable, and adaptive solutions have positioned Intelligent Transportation Systems (ITS) at the forefront of modern infrastructure innovation. At the core of ITS lies the challenge of autonomous decision-making across dynamic, large scale, and uncertain environments where multiple agents traffic signals, autonomous vehicles, or fleet units must coordinate effectively. Multi Agent Reinforcement Learning (MARL) offers a promising paradigm for addressing these challenges by enabling distributed agents to jointly learn optimal strategies that balance individual objectives with system wide efficiency. This paper presents a comprehensive survey of MARL applications in ITS. We introduce a structured taxonomy that categorizes MARL approaches according to coordination models and learning algorithms, spanning value based, policy based, actor critic, and communication enhanced frameworks. Applications are reviewed across key ITS domains, including traffic signal control, connected and autonomous vehicle coordination, logistics optimization, and mobility on demand systems. Furthermore, we highlight widely used simulation platforms such as SUMO, CARLA, and CityFlow that support MARL experimentation, along with emerging benchmarks. The survey also identifies core challenges, including scalability, non stationarity, credit assignment, communication constraints, and the sim to real transfer gap, which continue to hinder real world deployment.
\end{abstract}




\maketitle

\section{INTRODUCTION}\label{introduction}
The global evolution of urban mobility is marked by growing transportation demands, increasing urban congestion, and the pressing need for sustainable and efficient mobility solutions \cite{karjalainen2021urban, serdar2022urban}. To meet these challenges, Intelligent Transportation Systems (ITS) have emerged as a cornerstone of modern infrastructure, aiming to integrate advanced sensing, control, and communication technologies to enhance traffic efficiency, safety, and environmental performance \cite{cress2023intelligent}.

At the heart of ITS lies a critical need for autonomous decision-making in complex, dynamic, and often uncertain environments \cite{azadani2021driving}. Traditional rule-based and optimization-based methods often fall short when faced with large-scale, stochastic, and multi-agent traffic scenarios \cite{azadani2021driving}. In this context, Reinforcement Learning (RL) has gained traction as a powerful data-driven control paradigm capable of learning optimal or near-optimal policies through interaction with the environment \cite{nama2021machine}. However, real-world transportation systems are rarely single-agent systems \cite{wei2021recent}. Instead, they involve numerous distributed and interacting agents such as traffic lights, autonomous vehicles, or fleet units making Multi-Agent Reinforcement Learning (MARL) particularly relevant \cite{lin2021deep}.

While standard reinforcement learning (RL) has shown success in isolated tasks, multi-agent reinforcement learning (MARL) uniquely enables agents to learn both individual policies and coordination strategies, facilitating cooperative traffic signal optimization across large networks, vehicle-to-vehicle (V2V) and vehicle-to-infrastructure (V2I) coordination in connected autonomous driving, and scalable solutions for complex logistics, ride-sharing, and mobility-on-demand systems. However, despite its growing adoption, the MARL landscape in transportation remains fragmented, characterized by diverse algorithmic designs, inconsistent evaluation standards, and varied assumptions regarding agent interactions and reward structures. A consolidated understanding is urgently needed to clarify which methods are effective, under what conditions, and for what types of transportation problems.

\subsection{Scope and Contributions}
This paper provides a comprehensive survey of Multi-Agent Reinforcement Learning (MARL) approaches applied to Intelligent Transportation Systems (ITS), targeting both the transportation and artificial intelligence communities engaged in multi-agent challenges. A structured taxonomy is introduced to classify MARL architectures based on coordination models and learning algorithms. The survey offers a detailed analysis of MARL applications across various ITS domains, including traffic signal control and autonomous driving. In addition, commonly used simulation platforms and open-source benchmarks for MARL evaluation in ITS are reviewed. Key challenges such as scalability, safety, non-stationarity, and the sim-to-real transfer problem are identified as major barriers to practical deployment. The paper concludes by outlining future research directions, emphasizing opportunities in federate

\subsection{Organization of the Paper}
The remainder of the paper is organized as follows:
Section \ref{reinforcment_learning} introduces the fundamentals of reinforcement learning and its extension to multi-agent systems. Section \ref{marl_architecture} classifies MARL architectures and learning paradigms relevant to ITS. Section \ref{applications} reviews real-world applications of MARL across various ITS domains. Section \ref{challenges} presents core challenges and limitations of MARL in transportation. Section \ref{future} outlines future research directions, followed by conclusions in Section \ref{conclusion}.

\section{REINFORCEMENT LEARNING} \label{reinforcment_learning}
\subsection{Single Agent Reinforcement Learning (RL)}

\begin{figure}
  \centering
  \includegraphics[width=0.3\textwidth]{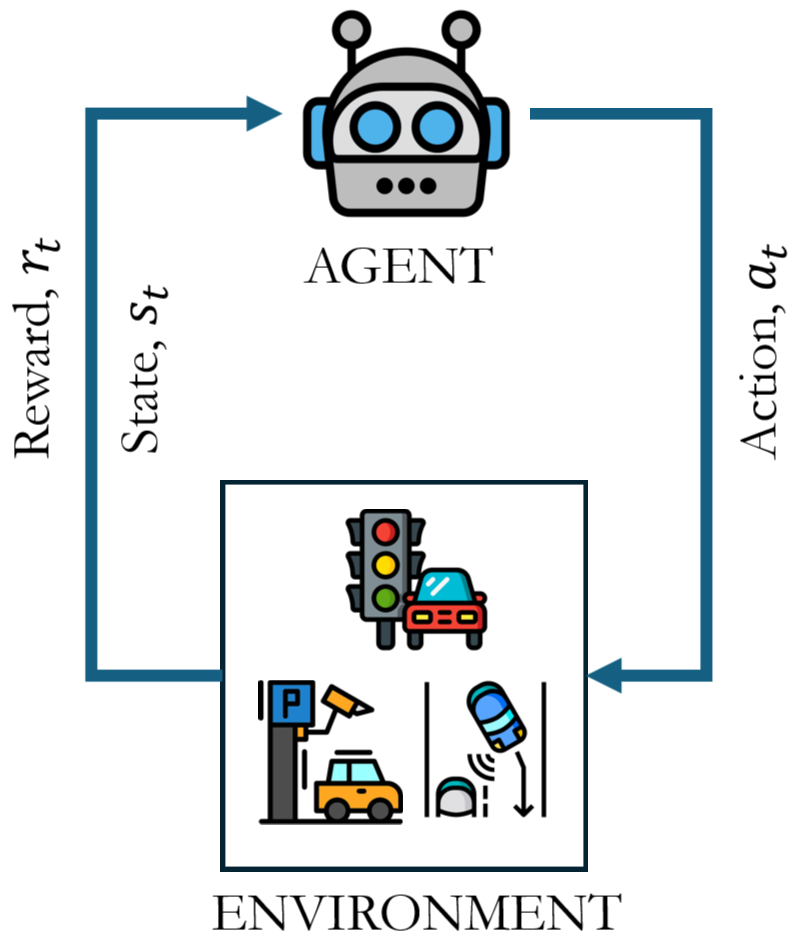}
    \caption{Illustration of the reinforcement learning loop: the agent interacts with the environment by taking actions $a_t$, and in return receives the next state $s_{t+1}$ and reward $r_t$, forming a continuous feedback cycle for learning optimal behavior.} \label{fig:rl_loop}
\end{figure}

Reinforcement Learning (RL) is a framework in which an agent learns to make decisions by interacting with an environment \cite{shakya2023reinforcement, matsuo2022deep, abel2023definition}. In the case of single agent reinforcement learning, there exists only one agent that perceives the environment's state, takes actions, and learns from the feedback it receives \cite{okafor2021heuristic, adetifa2023deep}. This setup forms the foundational structure of many RL algorithms and is depicted in Figure~\ref{fig:rl_loop}.

The agent-environment interaction is typically modeled as a Markov Decision Process (MDP), defined by a tuple $(S, A, P, R, \gamma)$ where \cite{bennett2024proximal, ororbia2022design, ren2022competitive}:
\begin{itemize}
    \item $S$ is the set of possible states,
    \item $A$ is the set of possible actions,
    \item $P(s'|s, a)$ defines the transition probabilities,
    \item $R(s, a)$ is the reward function,
    \item $\gamma \in [0,1)$ is the discount factor.
\end{itemize}

The agent is a computational unit responsible for selecting actions based on its current policy $\pi(a|s)$ \cite{he2023nearly}. At each time step $t$, the agent receives a state $s_t$ from the environment and selects an action $a_t$ \cite{chala2025mathematical}. This action is executed in the environment, which responds by providing a scalar reward $r_t$ and the next state $s_{t+1}$ \cite{ronca2022markov}.

The cycle continues as the agent updates its policy or value estimates using this feedback \cite{kurniawati2022partially}. The aim is to maximize the cumulative reward over time, often formalized as the return \cite{kumar2023policy}:
\begin{equation}
    G_t = \sum_{k=0}^\infty \gamma^k R_{t+k+1}
\end{equation}

The environment in Figure~\ref{fig:rl_loop} is represented with various real-world elements such as traffic signals, parking, and sensors illustrating that reinforcement learning can be applied to complex domains like autonomous driving, smart traffic control, or robotic systems.

Single agent reinforcement learning assumes that the environment is stationary and non-adversarial. The learning process involves trial and error, where the agent explores different strategies and improves its behavior based on observed outcomes \cite{abel2023definition}. Over time, the agent converges to an optimal or near-optimal policy that maximizes its expected return \cite{yu2025dapo}. This simple yet powerful interaction loop serves as the basis for more advanced scenarios in multi-agent systems, partially observable environments, and continuous control tasks.

\begin{figure}
  \centering
  \includegraphics[width=0.4\textwidth]{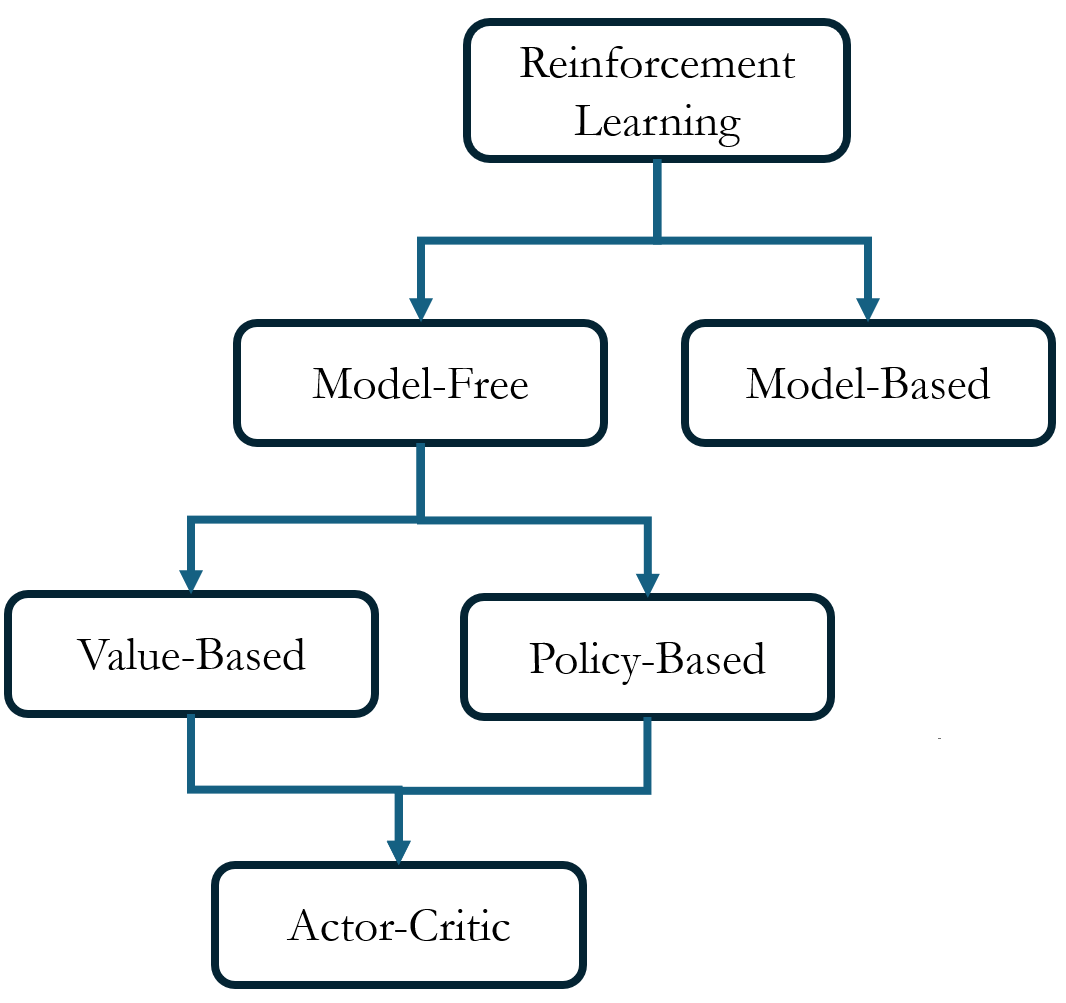}
    \caption{Hierarchy of Reinforcement Learning Methods: Categorizing Approaches into Model-Free and Model-Based} \label{fig:learning_algorithms}
\end{figure}

Reinforcement Learning (RL) algorithms are commonly classified into three main categories based on their learning strategies: value-based, policy-based, and actor-critic methods, as illustrated in Figure \ref{fig:learning_algorithms}. This categorization highlights the different ways in which each approach models the agent's decision-making process during interaction with the environment. The following sections provide an overview of these learning strategies.
\\

\subsubsection{Value-Based Methods}
Value-based reinforcement learning (RL) methods aim to learn a policy that maximizes the expected cumulative reward by estimating the value of states or state-action pairs \cite{mckenzie2022modern}. In such methods, the agent interacts with the environment and updates its internal estimates of the desirability of being in particular states or performing specific actions \cite{byeon2023advances, boute2022deep}. These methods do not require an explicit model of the environment and are widely used due to their simplicity and general applicability \cite{esteso2023reinforcement}.

The central idea in value-based approaches is the notion of a value function, which quantifies the expected return from a state or from taking an action in a state under a specific policy \cite{liu2021sharp}. The state-value function under a policy $\pi$, denoted as $V^\pi(s)$, represents the expected total discounted reward when starting in state $s$ and following the policy $\pi$ thereafter \cite{moradimaryamnegari2022model}. It is formally defined as:

\begin{equation}
    V^\pi(s) = \mathbb{E}_\pi \left[ \sum_{k=0}^\infty \gamma^k R_{t+k} \mid S_t = s \right]
\end{equation}

where $\gamma \in [0, 1)$ is the discount factor, which determines the present value of future rewards, and $R_{t+k}$ is the reward received $k$ steps into the future \cite{okafor2021photovoltaic}.

Similarly, the action-value function $Q^\pi(s, a)$ gives the expected return when the agent starts from state $s$, takes action $a$, and then follows the policy $\pi$:

\begin{equation}
    Q^\pi(s, a) = \mathbb{E}_\pi \left[ \sum_{k=0}^\infty \gamma^k R_{t+k} \mid S_t = s, A_t = a \right]
\end{equation}

The goal in reinforcement learning is to find an optimal policy $\pi^*$ that yields the highest expected return from each state \cite{ball2023efficient}. The corresponding optimal state-value function is defined as:

\begin{equation}
    V^*(s) = \max_\pi V^\pi(s)
\end{equation}

and the optimal action-value function as:

\begin{equation}
    Q^*(s, a) = \max_\pi Q^\pi(s, a)
\end{equation}

From the optimal action-value function, an optimal policy can be derived by selecting the action that maximizes the expected return:

\begin{equation}
    \pi^*(s) = \arg\max_{a} Q^*(s, a)
\end{equation}

To compute these value functions, recursive relationships known as Bellman equations are employed \cite{ball2023efficient}. The Bellman expectation equation for the state-value function under policy $\pi$ is given by:

\begin{equation}
    V^\pi(s) = \sum_{a} \pi(a|s) \sum_{s', r} P(s', r | s, a) \left[ r + \gamma V^\pi(s') \right]
\end{equation}

where $P(s', r | s, a)$ is the probability of transitioning to state $s'$ and receiving reward $r$ after taking action $a$ in state $s$ \cite{luo2024survey}.

Similarly, the Bellman expectation equation for the action-value function is:

\begin{equation}
    Q^\pi(s,a) = \sum_{s', r} P(s', r | s, a) \left[ r + \gamma \sum_{a'} \pi(a'|s') Q^\pi(s', a') \right]
\end{equation}

When seeking the optimal value functions, the Bellman optimality equations replace the expectations over the policy with maximizations \cite{kaufmann2023survey}. The optimal state-value function satisfies:

\begin{equation}
    V^*(s) = \max_{a} \sum_{s', r} P(s', r | s, a) \left[ r + \gamma V^*(s') \right]
\end{equation}

and the optimal action-value function is defined recursively as:

\begin{equation}
    Q^*(s,a) = \sum_{s', r} P(s', r | s, a) \left[ r + \gamma \max_{a'} Q^*(s', a') \right]
\end{equation}

In practice, these value functions are typically estimated through interaction with the environment. One popular approach is temporal-difference (TD) learning, which updates value estimates using bootstrapping \cite{paniri2021ant}. For instance, the TD(0) update rule for the state-value function is:

\begin{equation}
    V(s_t) \leftarrow V(s_t) + \alpha \left[ R_{t+1} + \gamma V(s_{t+1}) - V(s_t) \right]
\end{equation}

where $\alpha$ is the learning rate.

For estimating the action-value function, two widely used algorithms are Q-learning and SARSA. Q-learning is an off-policy method that learns the optimal value function regardless of the agent's current policy. Its update rule is:

\begin{equation}
    Q(s_t, a_t) \leftarrow Q(s_t, a_t) + \alpha \left[ R_{t+1} + \gamma \max_{a'} Q(s_{t+1}, a') - Q(s_t, a_t) \right]
\end{equation}

SARSA (State-Action-Reward-State-Action), on the other hand, is an on-policy method that updates the value function based on the agent’s actual behavior:

\begin{equation}
    Q(s_t, a_t) \leftarrow Q(s_t, a_t) + \alpha \left[ R_{t+1} + \gamma Q(s_{t+1}, a_{t+1}) - Q(s_t, a_t) \right]
\end{equation}

These value-based reinforcement learning methods are foundational in the field and serve as the basis for more advanced techniques, including deep reinforcement learning. By learning to accurately estimate value functions, agents can make increasingly effective decisions in complex, uncertain environments.
\\

\subsubsection{Policy-Based Methods}
Policy-based reinforcement learning methods take a different approach from value-based methods by directly parameterizing and optimizing the policy itself, rather than deriving it indirectly from value functions \cite{yang2023hybrid}. These methods are particularly well-suited to environments with large or continuous action spaces, where maintaining and maximizing action-value functions becomes computationally expensive or unstable \cite{nousiainen2022toward}.

In policy-based methods, the agent's behavior is described by a policy $\pi(a|s;\theta)$, which defines the probability of selecting action $a$ in state $s$, given a set of parameters $\theta$ \cite{wang2022diffusion}. The objective is to find the optimal policy parameters $\theta^*$ that maximize the expected return from each state \cite{lopez2023efficient}.

The performance of a policy is typically quantified using the objective function $J(\theta)$, which measures the expected cumulative reward when following the policy $\pi_\theta$ \cite{wang2022diffusion}:

\begin{equation}
    J(\theta) = \mathbb{E}_{\pi_\theta} \left[ \sum_{t=0}^\infty \gamma^t R_t \right]
\end{equation}

To optimize this objective, gradient ascent is applied. The core idea is to update the parameters $\theta$ in the direction of the gradient of $J(\theta)$ with respect to $\theta$ \cite{hambly2023recent}. The update rule is:

\begin{equation}
    \theta \leftarrow \theta + \alpha \nabla_\theta J(\theta)
\end{equation}

The fundamental result enabling this update is the policy gradient theorem, which provides a way to compute the gradient of the expected return without needing to differentiate through the state transition probabilities \cite{guo2022real}. The policy gradient is given by:

\begin{equation}
    \nabla_\theta J(\theta) = \mathbb{E}_{\pi_\theta} \left[ \nabla_\theta \log \pi_\theta(a|s) \cdot Q^{\pi_\theta}(s,a) \right]
\end{equation}

In practice, since the true action-value function $Q^{\pi_\theta}(s, a)$ is usually unknown, various estimators are used \cite{wang2021review}. One common approach is to use the return $G_t$ observed from a trajectory:

\begin{equation}
    \nabla_\theta J(\theta) \approx \mathbb{E}_{\pi_\theta} \left[ \nabla_\theta \log \pi_\theta(a_t|s_t) \cdot G_t \right]
\end{equation}

This forms the basis of the REINFORCE algorithm, a Monte Carlo policy gradient method. While simple and unbiased, REINFORCE suffers from high variance \cite{zamfirache2022policy}. To reduce this variance, a baseline function $b(s_t)$ often chosen as the state-value function $V^{\pi}(s_t)$ can be subtracted without introducing bias:

\begin{equation}
    \nabla_\theta J(\theta) \approx \mathbb{E}_{\pi_\theta} \left[ \nabla_\theta \log \pi_\theta(a_t|s_t) \cdot (G_t - b(s_t)) \right]
\end{equation}

Using the state-value function as a baseline leads to the advantage function:

\begin{equation}
    A^\pi(s,a) = Q^\pi(s,a) - V^\pi(s)
\end{equation}

In this case, the policy gradient becomes:

\begin{equation}
    \nabla_\theta J(\theta) = \mathbb{E}_{\pi_\theta} \left[ \nabla_\theta \log \pi_\theta(a|s) \cdot A^\pi(s,a) \right]
\end{equation}

Policy-based methods naturally support stochastic policies, which are essential in partially observable or multi-agent environments \cite{kuba2021trust}. Another important property of policy-based methods is their ability to represent deterministic or continuous action distributions, which is difficult for value-based approaches \cite{zhou2021novel}. This makes policy gradient methods suitable for high-dimensional control tasks such as robotics and continuous control benchmarks \cite{liu2021deep, zhu2021deep}.

Despite their advantages, policy-based methods can suffer from problems such as slow convergence and sensitivity to hyperparameters \cite{hua2021learning}. As a result, much research has been devoted to improving their stability and efficiency, including algorithms such as Trust Region Policy Optimization (TRPO), Proximal Policy Optimization (PPO), and Soft Actor-Critic (SAC) \cite{kuba2021trust}.
\\

\subsubsection{Actor Critic Methods}
\begin{figure}
  \centering
  \includegraphics[width=0.4\textwidth]{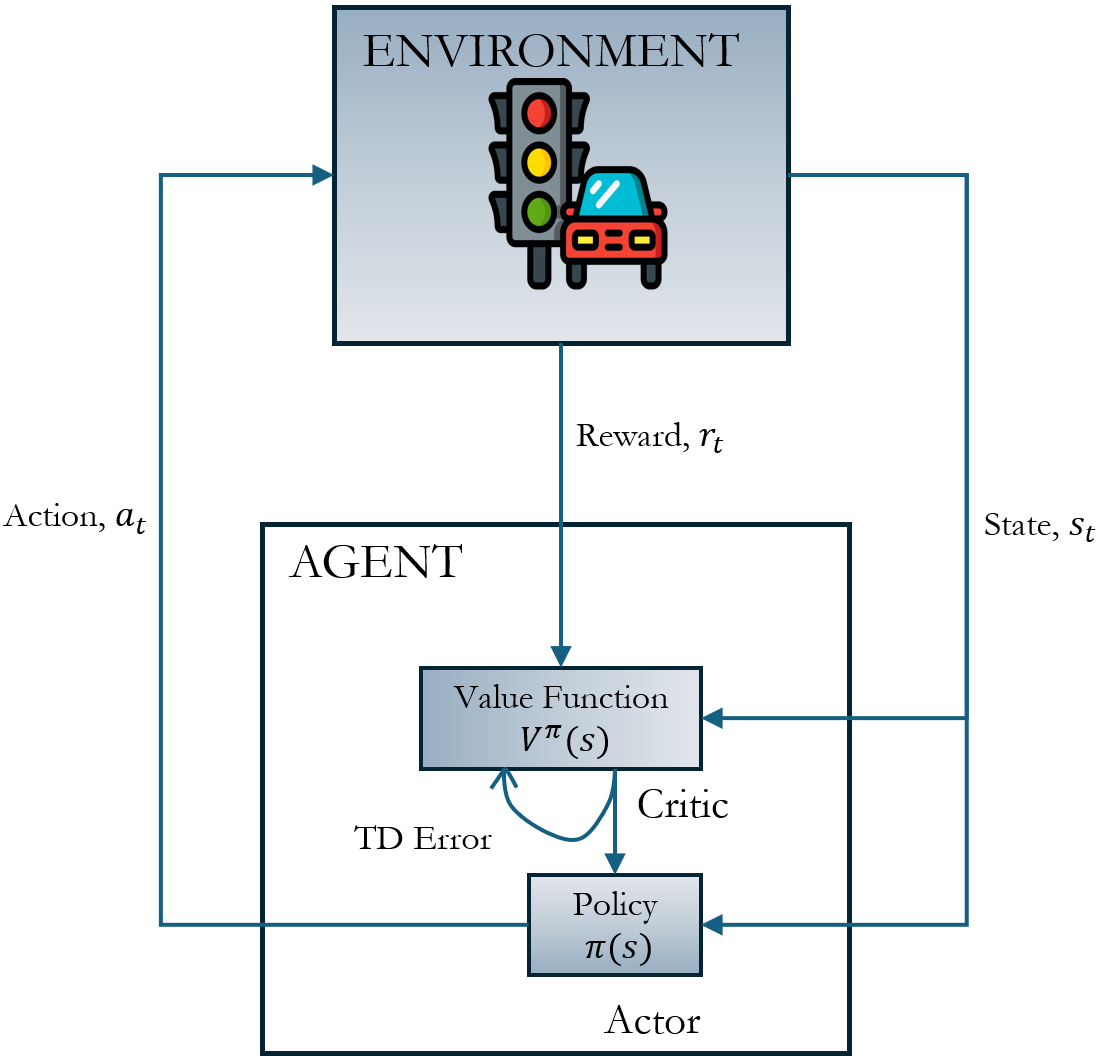}
    \caption{Illustration of the Actor-Critic Reinforcement Learning Framework} \label{fig:actor-critic}
\end{figure}

Actor-critic methods combine the key principles of value-based and policy-based reinforcement learning approaches \cite{sun2022high}. In these methods visualized in Figure \ref{fig:actor-critic}, the agent maintains two separate models: an actor, which is responsible for selecting actions according to a policy, and a critic, which evaluates the chosen actions by estimating value functions \cite{zamfirache2023neural}. This architecture enables the agent to improve its decision-making process through a blend of policy optimization and value estimation \cite{chen2022adaptive}.

The actor corresponds to the policy function $\pi_\theta(a|s)$, parameterized by $\theta$, and determines the agent's behavior by specifying the probability distribution over actions given the current state \cite{cheng2022adversarially}. The critic, on the other hand, estimates either the state-value function $V^{\pi}(s)$ or the action-value function $Q^{\pi}(s,a)$, using a separate set of parameters, often denoted by $\phi$ \cite{cheng2022adversarially}.

The core idea of actor-critic methods is to use the critic to provide a low-variance estimate of the policy gradient, which guides the actor's updates \cite{zamfirache2023q}. Instead of relying on high-variance returns from entire trajectories, the actor adjusts its policy parameters in the direction of the estimated advantage:

\begin{equation}
    \nabla_\theta J(\theta) \approx \mathbb{E}_{\pi_\theta} \left[ \nabla_\theta \log \pi_\theta(a_t|s_t) \cdot A^\pi(s_t, a_t) \right]
\end{equation}

Here, $A^\pi(s, a)$ is the advantage function, defined as:

\begin{equation}
    A^\pi(s, a) = Q^\pi(s, a) - V^\pi(s)
\end{equation}

The advantage quantifies how much better (or worse) an action is compared to the average action in that state under the current policy \cite{chen2021bringing}. A positive advantage indicates that the action yields a higher return than expected, encouraging the actor to increase the probability of selecting it.

In practice, $A^\pi(s, a)$ is often approximated using bootstrapped estimates. A common estimator is the one-step temporal difference advantage \cite{udekwe2024comparing}:

\begin{equation}
    \hat{A}_t = R_{t+1} + \gamma V(s_{t+1}) - V(s_t)
\end{equation}

This can be extended to multi-step or generalized advantage estimation (GAE) for improved stability and reduced variance. The critic itself is updated using standard temporal-difference learning rules. When estimating the state-value function, the critic's update is typically \cite{udekwe2024comparing, okafor2021photovoltaic, okafor2021solar, udekwe2025evaluating}:

\begin{equation}
    V(s_t) \leftarrow V(s_t) + \alpha \left[ R_{t+1} + \gamma V(s_{t+1}) - V(s_t) \right]
\end{equation}

When the action-value function is used instead, the update resembles that of Q-learning or SARSA, depending on whether the method is off-policy or on-policy.

Actor-critic methods can be further categorized based on how the policy is represented and updated \cite{chen2021bringing}. In discrete action spaces, the policy is usually stochastic, and the actor samples from $\pi_\theta(a|s)$. In continuous action spaces, deterministic policies are often used, and the deterministic policy gradient theorem provides the corresponding update \cite{zanette2021provable}:

\begin{equation}
    \nabla_\theta J(\theta) = \mathbb{E}_{s \sim \mathcal{D}} \left[ \nabla_\theta \pi_\theta(s) \cdot \nabla_a Q^\pi(s, a) \big|_{a=\pi_\theta(s)} \right]
\end{equation}

Several popular reinforcement learning algorithms are based on the actor-critic framework. These include Advantage Actor-Critic (A2C), Asynchronous Advantage Actor-Critic (A3C), Deep Deterministic Policy Gradient (DDPG), Twin Delayed Deep Deterministic Policy Gradient (TD3), Soft Actor-Critic (SAC), and Proximal Policy Optimization (PPO) \cite{zanette2021provable, yu2023actor}. Each of these algorithms introduces modifications to improve training stability, sample efficiency, or exploration.

Overall, actor-critic methods offer a powerful and flexible class of algorithms capable of handling high-dimensional and continuous control problems. By combining the strengths of policy gradients and value estimation, they enable effective learning in environments where purely value-based or policy-based methods may struggle.

\subsection{Multi-Agent Reinforcement Learning (MARL)}
Multi-Agent Reinforcement Learning (MARL) extends traditional reinforcement learning to environments involving multiple decision-making agents \cite{albrecht2024multi, gu2023safe}. Each agent interacts with a shared environment, learning to optimize its behavior based on received rewards and observed states . Unlike single-agent scenarios, MARL introduces additional complexity due to the presence of other learning agents, leading to non-stationary dynamics and coordination challenges \cite{li2021celebrating}.


\section{MARL ARCHITECTURES AND TAXONOMIES} \label{marl_architecture}
Multi-Agent Reinforcement Learning (MARL) involves multiple agents learning simultaneously within an environment, interacting with each other and the environment to achieve individual or shared objectives \cite{wen2022multi}. A fundamental challenge in MARL is coordination how agents align their policies or behaviors, especially under partial observability, non-stationarity, and sparse rewards \cite{huang2024multi}.

To handle these challenges, MARL research has introduced a variety of architectures and design taxonomies. One of the most crucial dimensions in this classification is the coordination model, which determines how agents share information, learn policies, and make decisions \cite{huang2024multi}. This section delves into three prominent coordination models which are shown in Figure \ref{fig_arc}:

\begin{figure*}
\centering
\subfloat[]{\includegraphics[width=0.3\textwidth]{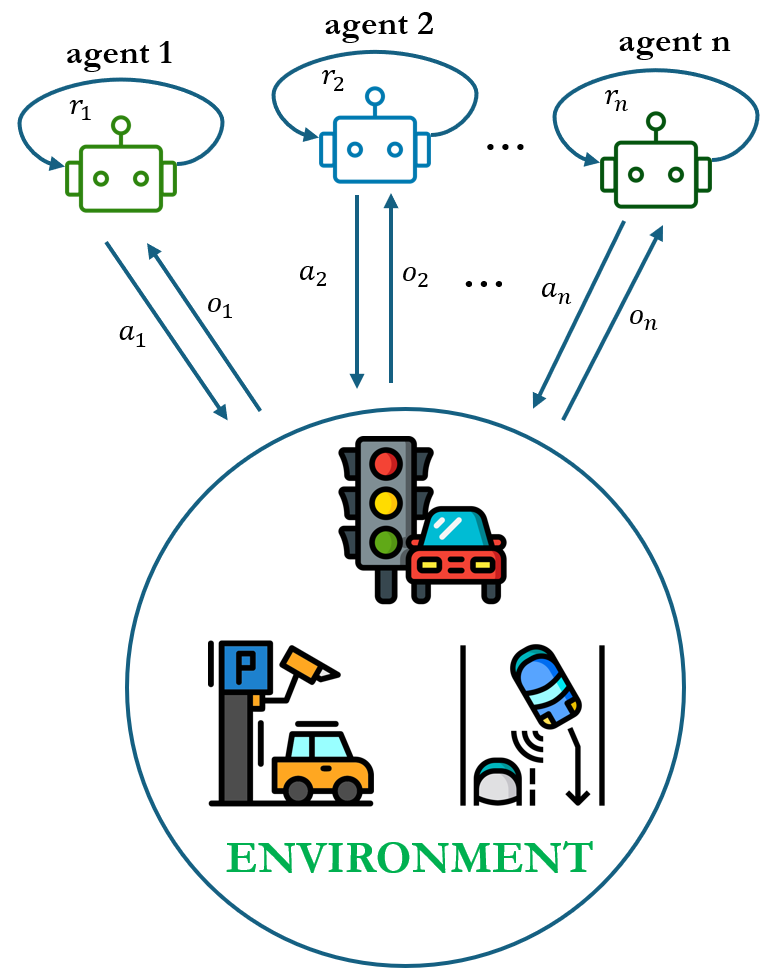}%
\label{fig:dtde}}
\hfil
\subfloat[]{\includegraphics[width=0.32\textwidth]{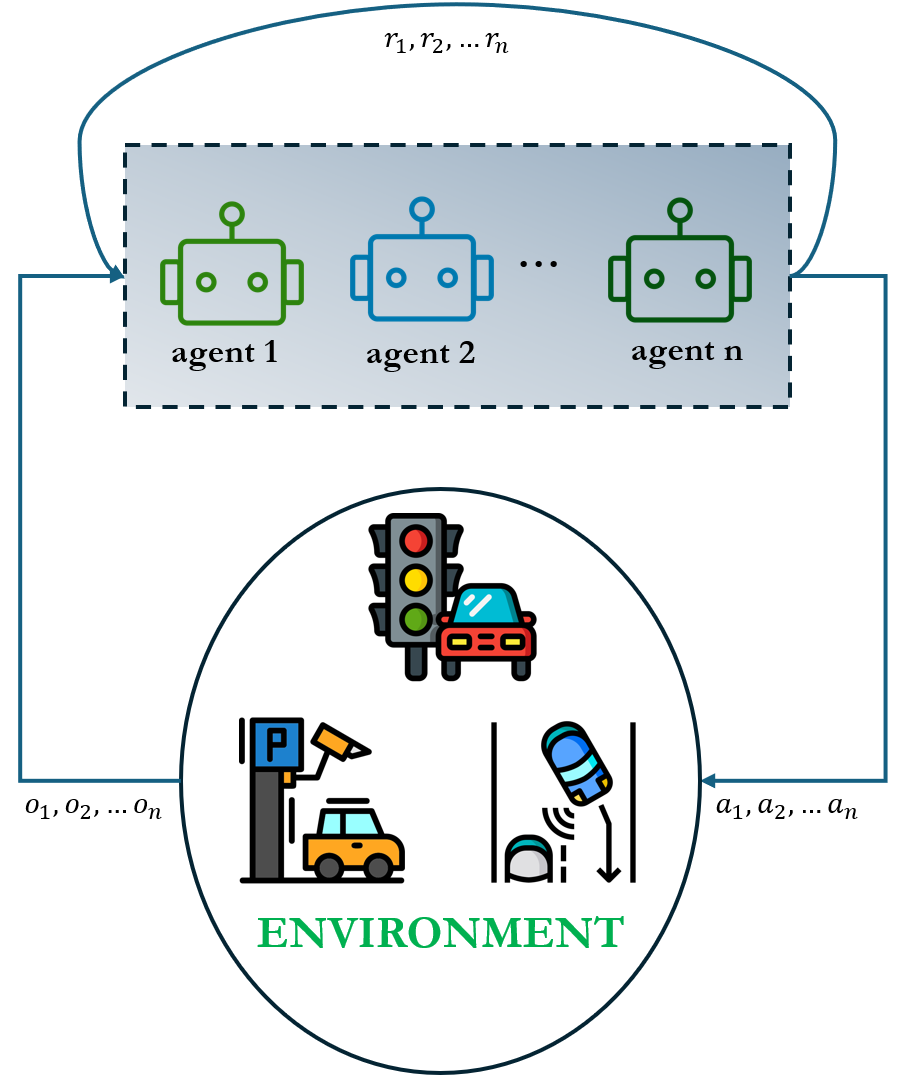}%
\label{fig:ctce}}
\hfil
\subfloat[]{\includegraphics[width=0.3\textwidth]{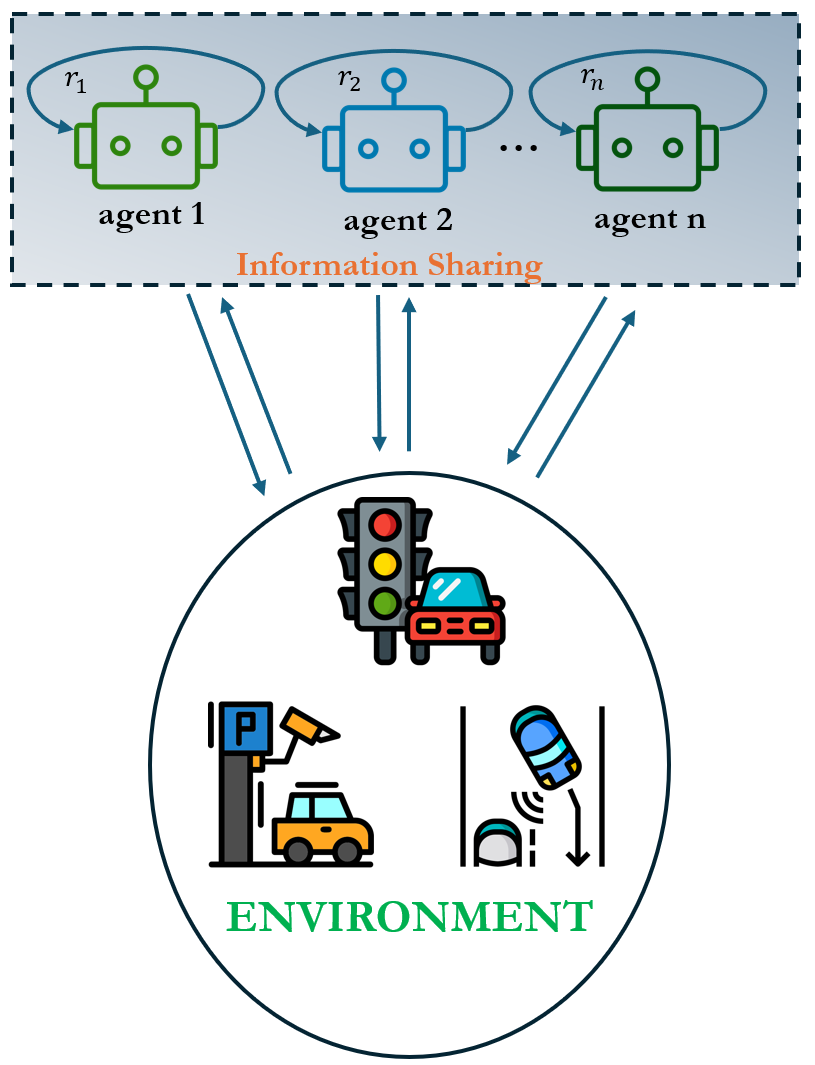}%
\label{fig:ctde}}
\caption{Multi-agent reinforcement learning architectures: (a) Decentralized Training and Decentralized Execution (DTDE)   agents learn independently with local observations and rewards; (b) Centralized Training with Centralized Execution (CTCE)   agents are trained and executed using shared global information; (c) Centralized Training with Decentralized Execution (CTDE)   agents are trained with global information but execute using only local observations.}
\label{fig_arc}
\end{figure*}

\subsection{Coordination Models}
In Multi-Agent Reinforcement Learning (MARL), coordination models determine how agents are trained and how they act during execution. A widely accepted way to categorize these models is based on the nature of training (centralized or decentralized) and execution (centralized or decentralized) \cite{kuba2021trust}. This framework gives rise to three principal coordination models: centralized training with centralized execution (CTCE), centralized training with decentralized execution (CTDE), and decentralized training with decentralized execution (DTDE) \cite{bettini2024benchmarl}.

\subsubsection{Centralized Training with Centralized Execution (CTCE)}
In centralized training with centralized execution (CTCE), both the learning and deployment phases rely on a centralized controller that possesses access to the complete global state, actions, and rewards of all agents \cite{ning2024survey}. The agents are treated as components of a single joint system, with policies often optimized together \cite{christianos2021scaling}. At execution time, this central authority continues to dictate the actions of all agents using the combined environmental information \cite{bae2022scientific}. CTCE enables optimal coordination and strategic joint behaviors, often achieving superior performance in fully observable and stable environments \cite{sun2024llm}. However, it is impractical in many real-world settings due to its high demands on communication, synchronization, and computational overhead \cite{huh2023multi}. Its reliance on continuous centralized control also renders it vulnerable to single points of failure, and it is largely restricted to simulation environments or highly controlled applications where latency and scalability are not critical concerns \cite{xia2021multi}.

\subsubsection{Centralized Training with Decentralized Execution (CTDE)}
Centralized training with decentralized execution (CTDE) is the most dominant coordination model in contemporary MARL research and applications \cite{xia2021multi}. In this framework, agents are trained in a centralized manner where they may share access to the global state, other agents' actions, or centralized critics that help stabilize learning and improve credit assignment \cite{huh2023multi}. However, once trained, the agents operate independently based solely on their local observations during execution \cite{ning2024survey}. This model achieves a balance between the advantages of centralized learning such as improved coordination, faster convergence, and better sample efficiency and the flexibility and scalability of decentralized action. CTDE is particularly useful in environments with partial observability and dynamic multi-agent interactions, such as in swarm robotics, autonomous vehicles, or multi-drone systems \cite{ning2024survey}. Despite its advantages, CTDE still depends on centralized infrastructure during training, which might be infeasible in fully distributed settings or environments where privacy and limited observability are critical  \cite{christianos2021scaling}.

\subsubsection{Decentralized Training with Decentralized Execution (DTDE)}
Decentralized training with decentralized execution (DTDE) represents the most decentralized coordination approach in MARL \cite{sun2024llm}. Here, each agent learns and acts independently using only its own local observations and rewards. There is no shared training infrastructure or global state, and each agent treats others as part of an evolving environment \cite{christianos2021scaling}. This model is simple, scalable, and naturally suited to highly distributed or communication-constrained systems, such as sensor networks or large-scale mobile ad hoc networks \cite{sun2024llm}. However, it suffers significantly from the non-stationarity of the environment since each agent's policy is changing in parallel with others \cite{christianos2021scaling}. This leads to unstable learning dynamics and often results in suboptimal policies, especially in cooperative tasks that require coordination. Moreover, the lack of shared information and explicit coordination mechanisms means that DTDE agents may converge to selfish or conflicting behaviors unless strong environmental incentives guide them toward cooperation\cite{sun2024llm}.

\subsection{MARL Algorithms}
The complexity of multi-agent environments marked by non-stationarity, partial observability, and inter-agent dependencies has driven the development of specialized algorithms that go beyond simple independent learning. These algorithms enhance stability, improve coordination, and allow for scalable learning across multiple agents. This section introduces several landmark algorithms in MARL used in intelligent transportation systems
\\

\subsubsection{\textbf{Value Decomposition Network (VDN)}}
Decomposes the joint action-value function into an additive sum of individual agent utilities \cite{wei2022vgn}. Suitable for cooperative settings under the CTDE paradigm \cite{wei2022vgn}.

  \begin{figure}
  \centering
  \includegraphics[width=0.4\textwidth]{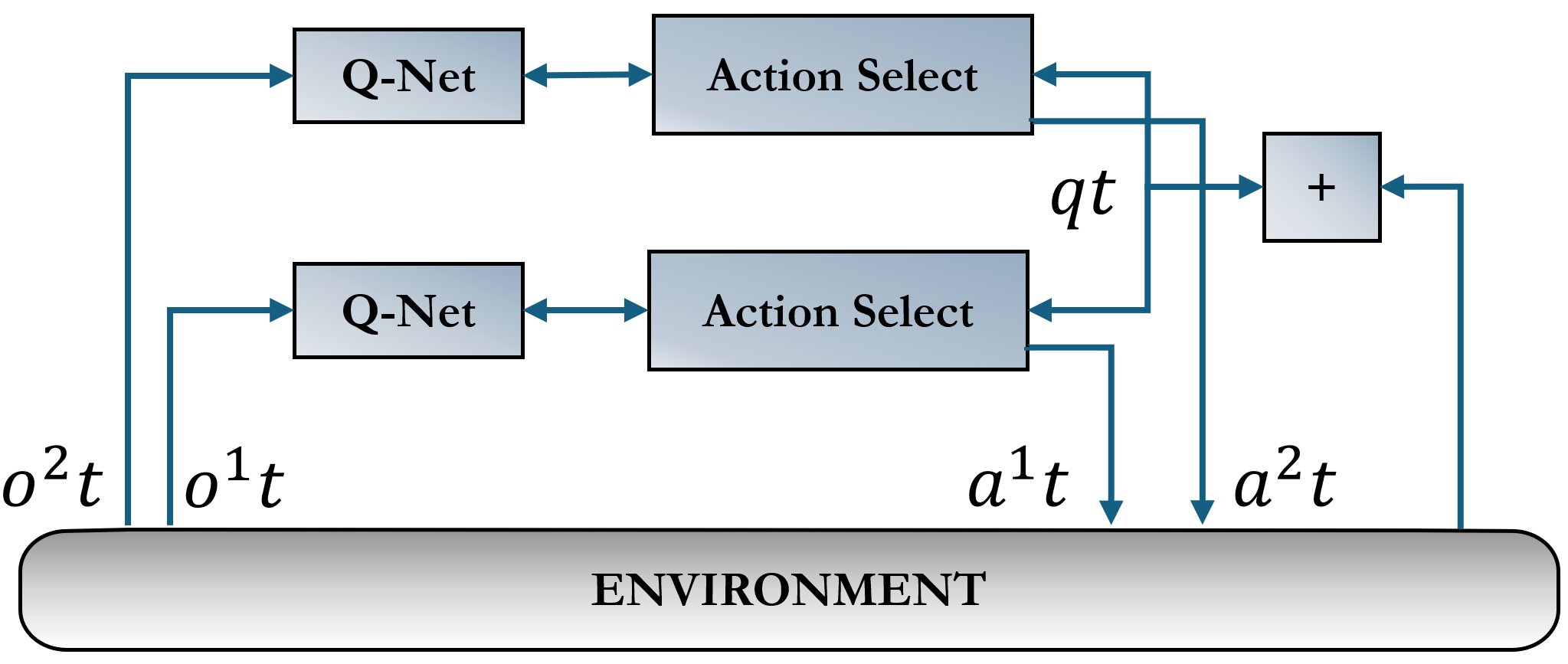}
    \caption{Illustration of the Value Decomposition Network (VDN) framework for multi-agent reinforcement learning. Each agent receives its own local observation ($o^1_t$, $o^2_t$) and uses an individual Q-network (Q-Net) to estimate its action-value function. Based on these estimates, actions ($a^1_t$, $a^2_t$) are selected. The corresponding Q-values are then aggregated (summed) to compute the joint action-value function $q_t$, which is used for centralized training.}
 \label{fig:vd}
\end{figure}

Figure \ref{fig:vd} illustrates the core architecture of the Value Decomposition Network (VDN), a foundational algorithm in cooperative Multi-Agent Reinforcement Learning (MARL) \cite{du2021survey}. VDN is designed for environments where multiple agents work together to maximize a shared reward \cite{wei2022vgn}. It leverages the concept of centralized training with decentralized execution (CTDE) by decomposing the global action-value function into individual agent components \cite{su2021value}.

At each time step $t$, every agent $i \in \{1, 2, \ldots, N\}$ receives a local observation $o^i_t$ from the environment \cite{fu2022revisiting}. These observations are processed independently by each agent's Q-network, producing an estimated action-value function $Q^i(o^i_t, a^i_t)$ for each possible action $a^i_t$ \cite{wei2022vgn}:

\[
Q^i_t = Q^i(o^i_t, a^i_t)
\]

Each agent selects its action $a^i_t$ using an action selection strategy such as $\epsilon$-greedy, based on its Q-values. These actions are executed simultaneously in the environment, resulting in a joint action vector \cite{fu2022revisiting}:

\[
\vec{a}_t = (a^1_t, a^2_t, \ldots, a^N_t)
\]

The environment then transitions to the next state and provides a shared reward $r_t$ to all agents \cite{fu2022revisiting}.

The core innovation of VDN lies in how it estimates the joint action-value function $Q_{\text{tot}}$ for the entire agent team. Instead of modeling the full joint Q-function directly which is computationally infeasible due to its exponential complexity VDN approximates it as a sum of individual Q-values \cite{wei2022vgn}:

\[
Q_{\text{tot}}(\vec{o}_t, \vec{a}_t) = \sum_{i=1}^{N} Q^i(o^i_t, a^i_t)
\]

This decomposition assumes independent contributions from each agent and allows the overall Q-learning update to be driven by the collective behavior \cite{du2021survey}:

\[
\theta \leftarrow \theta - \alpha \nabla_\theta \left( \sum_{i=1}^{N} Q^i(o^i_t, a^i_t) - y_t \right)^2
\]

where the target value is defined as\cite{du2021survey}:

\[
y_t = r_t + \gamma \max_{\vec{a}'} Q_{\text{tot}}(\vec{o}_{t+1}, \vec{a}')
\]

and $\gamma$ is the discount factor.

During execution, agents rely only on their local Q-networks and observations, ensuring decentralized policies while benefiting from centralized training using the global reward signal \cite{du2021survey}.
\\

\subsubsection{\textbf{QMIX}}
Extends VDN by allowing a monotonic mixing of individual Q-values using a mixing network. Enables more flexible coordination in cooperative tasks \cite{zhao2025qmix}.
  
\begin{figure}
  \centering
  \includegraphics[width=0.4\textwidth]{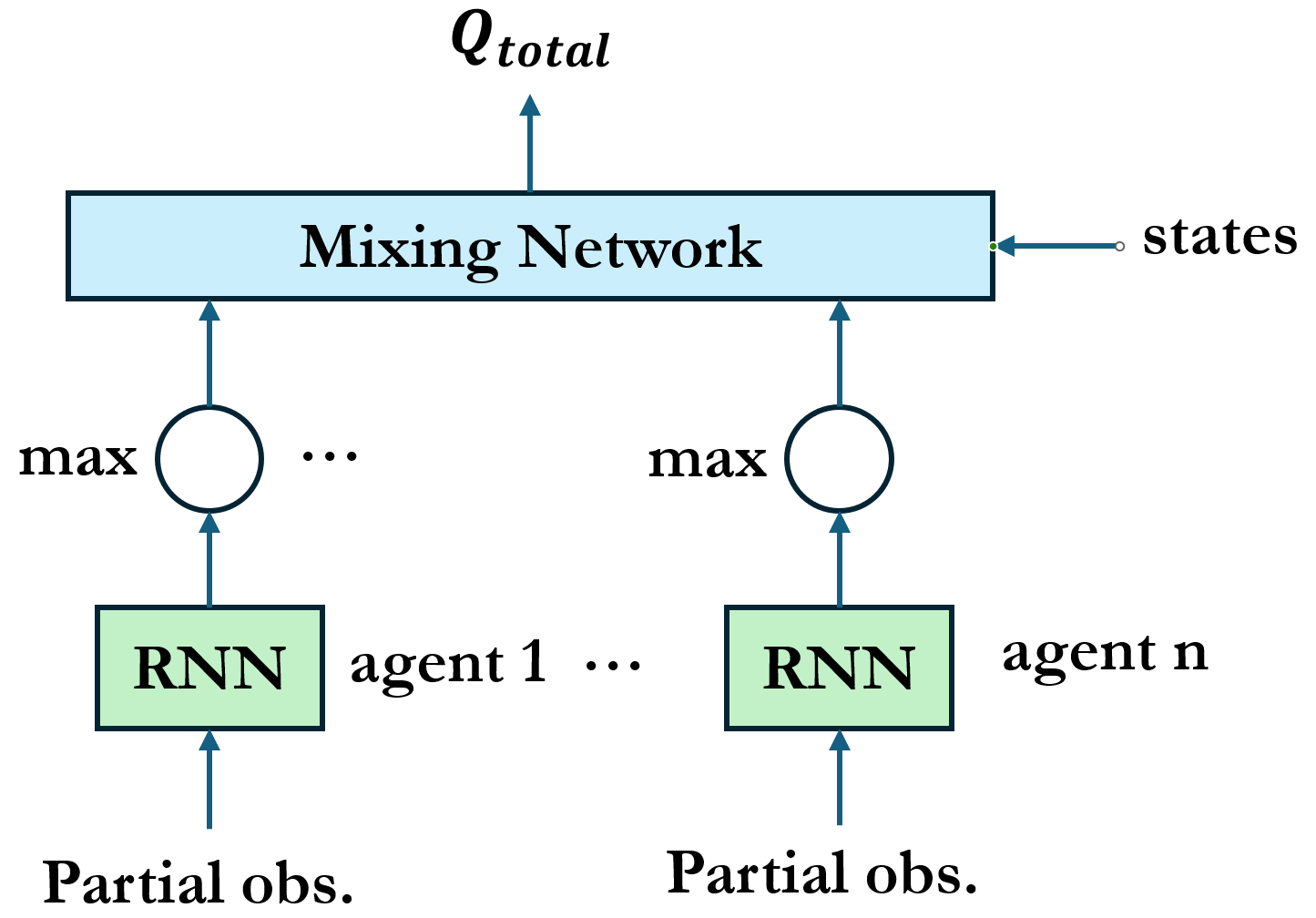}
    \caption{Illustration of the QMIX framework for value-based multi-agent reinforcement learning. Each agent receives a partial observation and uses a recurrent neural network (RNN) to estimate its individual action-value function. The selected Q-values are aggregated through a mixing network, which is conditioned on the global state and outputs a joint action-value function $Q_{\text{total}}$}
 \label{fig:qmix}
\end{figure}

Figure~\ref{fig:qmix} illustrates the architecture of QMIX, a value-based cooperative Multi-Agent Reinforcement Learning (MARL) algorithm designed to overcome the limitations of linear factorisation in VDN \cite{zhang2024sqix}. While VDN assumes the total team Q-value is the sum of individual agent Q-values, QMIX uses a more expressive \textit{nonlinear mixing network} that allows for flexible but still consistent value decomposition \cite{zhao2025qmix}.

Each agent $i \in \{1, 2, ..., N\}$ receives a \textit{partial observation} and encodes it using an agent-specific recurrent neural network (RNN). These RNNs output individual Q-values $Q^i(o^i_t, a^i_t)$ for local action-observation histories \cite{van2023applying}:

\[
Q^i_t = Q^i(o^i_{1:t}, a^i_t)
\]

The Q-values are combined via a centralized mixing network, which takes as input both \cite{wang2023qmix}:
\begin{itemize}
    \item the individual agent Q-values $Q^1_t, Q^2_t, ..., Q^N_t$, and
    \item the global state $s_t$ available during training.
\end{itemize}

This mixing network computes the joint action-value function $Q_{\text{tot}}$ \cite{heik2024application}:

\[
Q_{\text{tot}}(s_t, \vec{a}_t) = \text{MixingNet}(Q^1_t, Q^2_t, ..., Q^N_t; s_t)
\]

The key constraint in QMIX is monotonicity \cite{kim2024cooperative}:

\[
\frac{\partial Q_{\text{tot}}}{\partial Q^i} \geq 0, \quad \forall i
\]

This ensures that selecting actions to maximize each $Q^i$ also maximizes $Q_{\text{tot}}$, preserving decentralized execution while enabling a richer representational capacity during centralized training \cite{kim2024cooperative}.

The global reward is used to update the joint Q-function, and through backpropagation, each agent’s RNN is updated accordingly \cite{wang2023qmix}.

\begin{itemize}
    \item Execution: Fully decentralized   each agent acts based only on its own observation.
    \item Training: Centralized   full global state and joint Q-value are used.
\end{itemize}

This makes QMIX suitable for partially observable, cooperative tasks such as traffic signal control, robotic team coordination, and multi-drone exploration \cite{wang2023qmix}.
\\
\subsubsection{\textbf{Multi-Agent Deep Deterministic Policy Gradient (MADDPG)}}
Uses decentralized actors with centralized critics, enabling agents to handle mixed cooperative-competitive settings using continuous actions \cite{li2024multi}.

\begin{figure}
  \centering
  \includegraphics[width=0.4\textwidth]{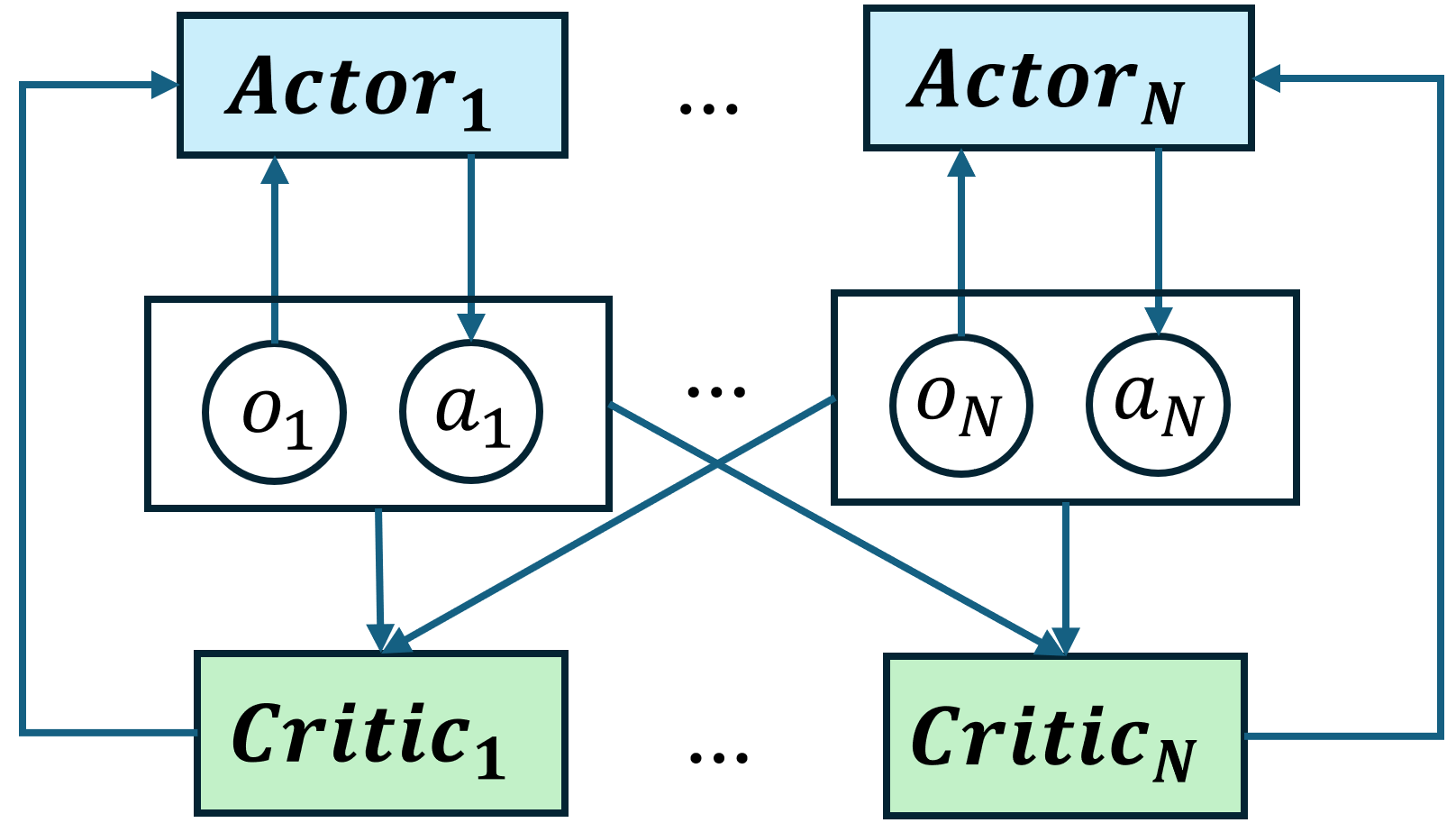}
    \caption{Illustration of the MADDPG (Multi-Agent Deep Deterministic Policy Gradient) architecture. Each agent $i$ has its own actor and critic networks. The actor receives the agent's local observation $o_i$ and outputs an action $a_i$, while the critic is trained with access to the joint observations and actions of all agents.}
 \label{fig:maddpg}
\end{figure}

Figure \ref{fig:maddpg} illustrates the architecture of MADDPG (Multi-Agent Deep Deterministic Policy Gradient), a prominent algorithm for learning policies in both cooperative and competitive multi-agent environments \cite{zhang2024vn}. MADDPG extends the Deep Deterministic Policy Gradient (DDPG) framework into the multi-agent setting by adopting a \textit{centralized training with decentralized execution (CTDE)} paradigm \cite{zhang2024vn}.

Each agent $i \in \{1, 2, ..., N\}$ is modeled as an actor–critic pair. The actor network  $\pi^i(o^i_t)$ outputs a deterministic action $a^i_t$ given the agent's partial observation $o^i_t$ \cite{yang2024multi}. During training, each agent's critic network $Q^i$ takes as input the joint observations and actions of all agents \cite{yang2024multi}:

\[
Q^i(o^1_t, ..., o^N_t, a^1_t, ..., a^N_t)
\]

This centralized critic allows each agent to account for the influence of other agents' actions during policy updates, which is critical in non-stationary multi-agent environments \cite{zakaryia2025task}.

The actor is updated by maximizing the critic's Q-value \cite{gao2021game}:

\[
\nabla_{\theta^i} J(\theta^i) = \mathbb{E}_{\mathbf{o}, \mathbf{a}} \left[ \nabla_{\theta^i} \pi^i(o^i_t) \nabla_{a^i} Q^i(o^1_t, ..., o^N_t, a^1_t, ..., a^N_t) \right]
\]

The critic is updated by minimizing the temporal-difference (TD) error \cite{zakaryia2025task}:

\[
L(\theta^i) = \mathbb{E}_{\mathbf{o}, \mathbf{a}, r, \mathbf{o'}} \left[ \left( Q^i(\mathbf{o}_t, \mathbf{a}_t) - y_t \right)^2 \right]
\]
\[
y_t = r^i_t + \gamma Q^{i'}(\mathbf{o}_{t+1}, \mathbf{a}_{t+1}')
\]

where $Q^{i'}$ and $\pi^{i'}$ are the target networks for stability \cite{zakaryia2025task}.

Execution Phase: After training, each agent uses only its local actor $\pi^i(o^i_t)$ to choose actions without requiring access to other agents' observations or actions   enabling fully decentralized execution \cite{zakaryia2025task}.

MADDPG has been effectively applied to:
\begin{itemize}
    \item Cooperative communication tasks
    \item Competitive games and navigation
    \item Autonomous vehicle interactions and adversarial driving
\end{itemize}

\subsubsection{\textbf{Multi-Agent Proximal Policy Optimization (MAPPO)}}
An extension of PPO to multi-agent domains with centralized critics and stable, scalable performance \cite{chen2021multiagent}.
\begin{figure}
  \centering
  \includegraphics[width=0.4\textwidth]{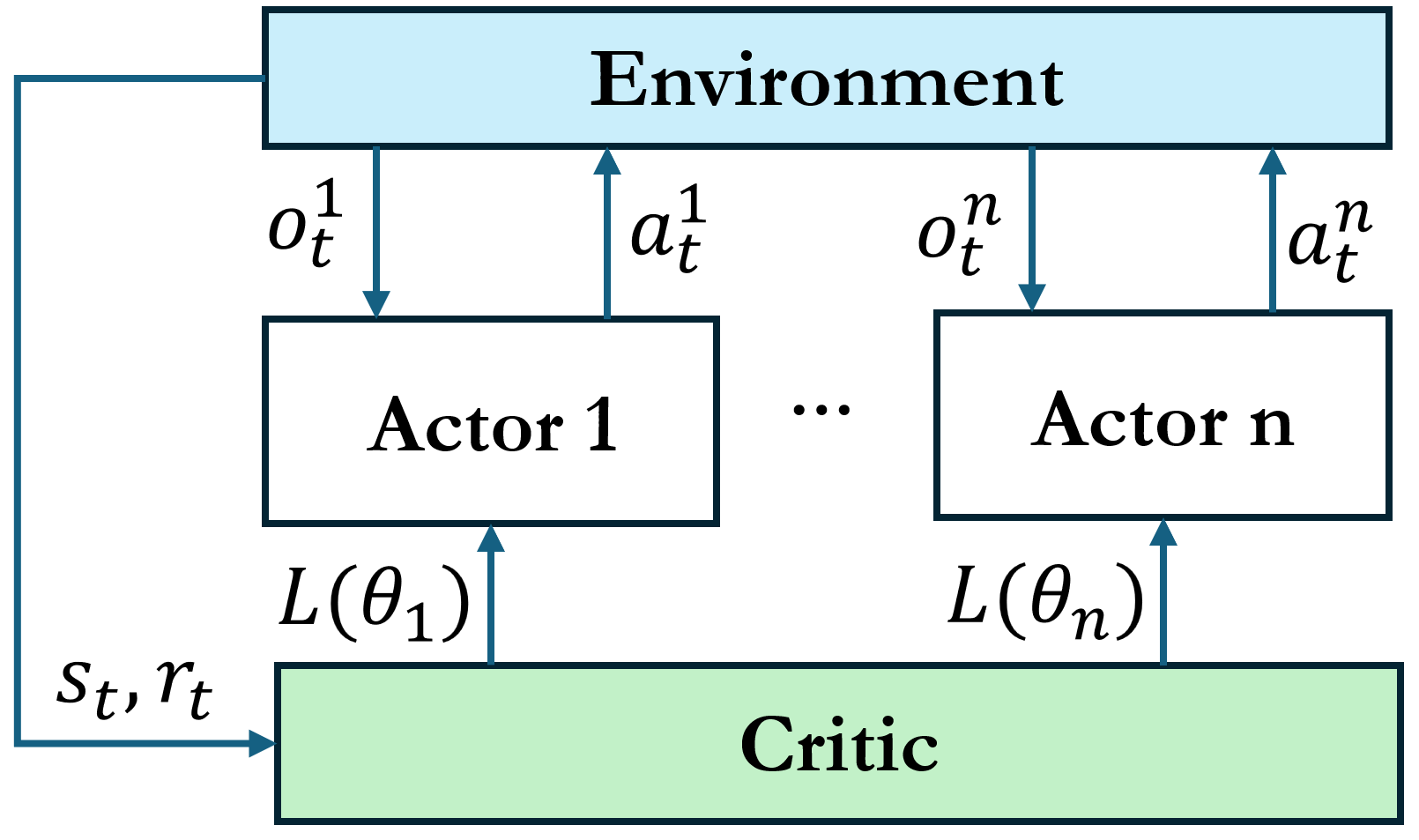}
   \caption{Illustration of the MAPPO (Multi-Agent Proximal Policy Optimization) architecture. Each agent receives a local observation ($o_t^1, \dots, o_t^n$) and uses its own actor to produce an action ($a_t^1, \dots, a_t^n$). A centralized critic receives the global state $s_t$ and reward $r_t$ to compute the policy loss $L(\theta_i)$ for updating each actor $i$. }
 \label{fig:mappo}
\end{figure}

Figure~\ref{fig:mappo} depicts the architecture of MAPPO (Multi-Agent Proximal Policy Optimization), an actor–critic MARL algorithm adapted from PPO \cite{cai2024joint}. It leverages centralized training with decentralized execution (CTDE), enabling scalable and stable learning in cooperative multi-agent environments.

Each agent $i \in \{1, 2, \ldots, n\}$ receives a local observation $o^i_t$ from the environment and outputs an action $a^i_t$ via its decentralized actor policy $\pi_{\theta_i}(a^i_t | o^i_t)$ \cite{cai2024joint}.

The centralized critic uses the global state $s_t$ and all agents’ actions to evaluate the joint policy and compute the shared or individual value function $V(s_t)$ or $Q(s_t, \vec{a}_t)$ \cite{cai2024joint}. The critic is used to compute the surrogate objective and advantage estimates required for PPO updates \cite{chen2021multiagent}.

Each actor is updated using the PPO clipped surrogate objective \cite{chen2021multiagent}:
\[
L(\theta_i) = \mathbb{E}_t \left[ \min \left( r_t(\theta_i) \hat{A}_t, \text{clip}(r_t(\theta_i), 1 - \epsilon, 1 + \epsilon) \hat{A}_t \right) \right]
\]
where $r_t(\theta_i) = \frac{\pi_{\theta_i}(a^i_t|o^i_t)}{\pi_{\theta_i^{\text{old}}}(a^i_t|o^i_t)}$ is the probability ratio, and $\hat{A}_t$ is the advantage function derived from the critic \cite{chen2021multiagent}.

Key properties of MAPPO:
\begin{itemize}
    \item Centralized critic during training enables credit assignment using global information.
    \item Decentralized actors ensure agents act based only on local observations.
    \item Clipped updates stabilize policy improvement, inherited from single-agent PPO.
\end{itemize}

MAPPO has been widely applied in:
\begin{itemize}
    \item Multi-robot coordination
    \item Multi-agent games (e.g., StarCraft II micromanagement)
    \item Urban traffic control and air traffic deconfliction
\end{itemize}
.
\\

\subsubsection{\textbf{Hysteretic Q-Learning}}
Hysteretic Q-learning is a value-based reinforcement learning algorithm that introduces asymmetry in learning rates to enhance stability in multi-agent settings \cite{brown2023deep}. It is particularly effective in cooperative environments where agents face non-stationarity and partial observability due to the presence of other learning agents \cite{brown2023deep}.

Unlike standard Q-learning, which applies a single learning rate for all updates, hysteretic Q-learning uses two distinct learning rates depending on whether the temporal-difference (TD) error is positive or negative \cite{brown2023deep}. The standard Q-learning update is given by \cite{brown2023deep}:

\[
Q(s_t, a_t) \leftarrow Q(s_t, a_t) + \alpha \left( r_t + \gamma \max_{a'} Q(s_{t+1}, a') - Q(s_t, a_t) \right)
\]

In hysteretic Q-learning, the update is modified as follows \cite{brown2023deep}:

\[
Q(s_t, a_t) \leftarrow Q(s_t, a_t) + \alpha^+ \delta_t \quad \text{if } \delta_t \geq 0
\]
\[
Q(s_t, a_t) \leftarrow Q(s_t, a_t) + \alpha^- \delta_t \quad \text{if } \delta_t < 0
\]

where $\delta_t$ is the TD-error, and $\alpha^+ > \alpha^-$, ensuring that the agent learns more readily from positive feedback while being cautious about penalizing actions that may appear suboptimal due to the exploratory or noisy behavior of other agents \cite{chalaki2021hysteretic}.

This hysteresis mechanism helps reduce instability caused by teammates’ fluctuating policies, especially in environments with shared team rewards or sparse feedback \cite{chalaki2021hysteretic}. It is often used in decentralized training settings, where each agent learns independently using only its local observations and rewards \cite{chalaki2021hysteretic}.

Applications of hysteretic Q-learning include:

\begin{itemize}
    \item cooperative robotic exploration
    \item decentralized traffic signal control
    \item coordinated navigation tasks
\end{itemize}
Hysteretic Q-learning serves as a foundational method in multi-agent reinforcement learning and has inspired further developments such as lenient Q-learning, which adds stochastic forgiveness to early mistakes, and Dec-HDRQN, which combines hysteresis with deep recurrent networks \cite{chalaki2021hysteretic}.
\\

\subsubsection{\textbf{Lenient Q-Learning}}
Lenient Q-learning is an extension of standard Q-learning tailored for cooperative multi-agent environments, particularly under stochastic dynamics and delayed rewards \cite{amhraoui2024expected}. It introduces the concept of leniency, allowing agents to be forgiving of early mistakes made by themselves or by teammates during exploration \cite{amhraoui2024expected}.

The main idea is to maintain a leniency value for each state–action pair that gradually decays over time \cite{amhraoui2024expected}. Initially, agents are optimistic about joint actions and ignore low rewards or penalties, enabling them to explore without prematurely discarding potentially beneficial actions due to the noisy influence of other agents \cite{amhraoui2024expected}.

The update rule modifies the Q-learning formula by incorporating a temperature-based leniency factor \cite{amhraoui2024expected}:

\[
Q(s_t, a_t) \leftarrow
\begin{cases}
Q(s_t, a_t) + \alpha\, \delta_t, & \text{if } \delta_t > 0 \text{ or } p_l \\
Q(s_t, a_t), & \text{else}
\end{cases}
\]
\vspace{-1mm}
\[
\delta_t = r_t + \gamma \max_{a'} Q(s_{t+1}, a') - Q(s_t, a_t)
\]

Each state–action pair $(s, a)$ is associated with a temperature $T(s, a)$ that decreases over time as the pair is visited more frequently \cite{amhraoui2024expected}. A Boltzmann-like function is used to compute the leniency $L(s, a)$ \cite{amhraoui2024expected}:

\[
L(s, a) = 1 - \exp\left(-\kappa \cdot T(s, a)\right)
\]

where $\kappa$ is a scaling factor. The agent uses $L(s, a)$ to probabilistically ignore negative updates during early exploration \cite{amhraoui2024expected}. This mechanism encourages optimism and helps the system converge to coordinated joint policies \cite{amhraoui2024expected}.

Key aspects of lenient Q-learning include:

\begin{itemize}
    \item tolerance of early suboptimal actions, improving learning in stochastic cooperative tasks
    \item decaying leniency to allow gradual enforcement of accurate value estimates
    \item decentralized execution, with each agent maintaining and updating its own Q-table and temperature values
\end{itemize}

Lenient Q-learning has been successfully applied in domains such as cooperative navigation, coordination games, and traffic signal control \cite{amhraoui2024expected}. It performs particularly well when optimal joint actions require synchronized behavior among agents, and where premature penalization can lead to policy divergence.
\\

\subsubsection{\textbf{Parameter Sharing Trust Region Policy Optimization (PS-TRPO)}}
PS-TRPO (Parameter Sharing Trust Region Policy Optimization) is a policy gradient-based multi-agent reinforcement learning approach that extends Trust Region Policy Optimization (TRPO) to cooperative settings by enforcing parameter sharing among agents \cite{wang2024urban}. It is particularly useful in environments where agents are homogeneous or perform similar roles \cite{wang2024urban}.

In PS-TRPO, a single policy network is shared among all agents. Each agent receives its own observation $o^i_t$ and acts independently according to a shared policy $\pi_{\theta}(a^i_t | o^i_t)$ \cite{wang2024urban}. Despite this decentralized execution, training is centralized using the aggregated experience of all agents \cite{yu2022trust}.

The TRPO update is based on maximizing the expected advantage while constraining the KL divergence between the old and new policies \cite{yu2022trust} \cite{yu2022trust}:

\[
\max_{\theta} \; \mathbb{E}_{(o, a) \sim \pi_{\theta_{\text{old}}}} \left[ \frac{\pi_{\theta}(a|o)}{\pi_{\theta_{\text{old}}}(a|o)} \hat{A}^{\pi_{\theta_{\text{old}}}}(o, a) \right]
\]
\[
\text{subject to } \mathbb{E}_{o} \left[ D_{\mathrm{KL}} \left( \pi_{\theta_{\text{old}}}(\cdot|o) \| \pi_{\theta}(\cdot|o) \right) \right] \leq \delta
\]

Here, $\hat{A}^{\pi_{\theta_{\text{old}}}}$ is the advantage estimate, and $\delta$ is a small trust region threshold that limits policy updates to stay within a reliable improvement region \cite{yu2022trust}.

Key characteristics of PS-TRPO include:

\begin{itemize}
    \item parameter sharing reduces model complexity and improves sample efficiency
    \item centralized training benefits from the collective experiences of all agents
    \item decentralized execution allows each agent to act independently in real time
\end{itemize}

PS-TRPO has been shown to perform well in tasks such as cooperative navigation, predator–prey games, and formation control, especially where symmetry among agents makes parameter sharing natural \cite{yu2022trust}. It also serves as a basis for more complex multi-agent policy gradient methods, including MAPPO and HAPPO.
\\

\subsubsection{\textbf{CommNet: Communication Network}}
CommNet (Communication Network) is a neural network architecture proposed for deep multi-agent reinforcement learning, in which agents are trained end-to-end with differentiable inter-agent communication \cite{khan2023communication}. Unlike traditional decentralized methods, CommNet enables agents to share information via continuous vectors during training and execution, allowing for learned coordination in cooperative tasks \cite{khan2023communication, du2021learning}.

In CommNet, each agent $i$ receives a local observation $o^i_t$ and processes it through an encoder to produce a hidden state $h^i_t$ \cite{du2021learning}. These hidden states are then averaged to produce a shared communication vector $c^i_t$ \cite{du2021learning}:

\[
c^i_t = \frac{1}{N-1} \sum_{j \neq i} h^j_t
\]

Each agent then updates its own hidden state using both its local information and the communication vector \cite{du2021learning}:

\[
h^i_{t+1} = f(h^i_t, c^i_t)
\]

The updated hidden state is then used to select an action $a^i_t$ via a policy network \cite{du2021learning}. The entire system is differentiable, and the agents are trained jointly using backpropagation and policy gradient methods such as REINFORCE or actor–critic approaches \cite{du2021learning}.

CommNet is especially effective in settings where:

\begin{itemize}
    \item agents require tight coordination (e.g., formation flying, cooperative navigation)
    \item partial observability limits individual performance
    \item communication can improve joint value estimation
\end{itemize}

Because the communication mechanism is fully differentiable and learned jointly with the policy, CommNet provides a natural and scalable way to integrate communication into MARL \cite{khan2023communication}. It is typically trained with centralized learning and executed in a decentralized manner, assuming agents can communicate their internal states in real time.

CommNet laid the foundation for subsequent communication-aware MARL methods, such as IC3Net, DIAL, and RIAL, which further explore gated, discrete, and attention-based communication mechanisms.

\begin{table*}[!t]
\centering
\caption{Summary of MARL Algorithms in Intelligent Transportation Systems}
\begin{tabular}{llp{2cm}p{7.8cm}}
\hline
\textbf{Algorithm} & \textbf{Agent Type} & \textbf{Structure} & \textbf{Features} \\
\hline
Hysteretic Q-Learning \cite{chalaki2021hysteretic} & Value-based & DTDE & Uses different learning rates for increasing and decreasing Q-values. Requires no communication between agents. \\
\\
Lenient Q-Learning \cite{mao2021multi} & Value-based & DTDE & Adds leniency to Q-updates by storing temperature values in experience replay, useful in stochastic environments. \\
\\
MAPPO \cite{tian2025optimal} & Policy Optimization & CTDE & Extends PPO to multi-agent settings with decentralized actors and a centralized critic. Supports stable learning through clipped surrogate objectives and trust region constraints. \\
\\
MADQN \cite{wang2024demand} & Value-based & DTDE & Uses importance sampling and low-dimensional encoding to handle multi-agent experience replay efficiently. \\
\\
PS-TRPO \cite{menda2018deep} & Policy Optimization & CTCE & Shares policy parameters during centralized training and updates them using trust-region constraints with curriculum learning. \\
\\
VDN \cite{wang2023variant} & Value-based & CTDE & Decomposes the joint Q-function into a sum of individual agent Q-values, enabling decentralized execution. \\
\\
QMIX \cite{heik2024application} & Value-based & CTDE & Extends VDN with a monotonic mixing network for better representational power while retaining decentralized execution. \\
\\
CommNet \cite{park2023multi} & Policy Optimization & CTDE & Learns communication and action policies jointly. Agents exchange continuous messages to select coordinated actions. \\
\\
MADDPG \cite{wu2020cooperative} & Actor-Critic & CTDE & Employs decentralized actors with centralized critics that observe all agent states and actions. Supports both cooperative and mixed settings with continuous action spaces. \\
\hline
\end{tabular}
\end{table*}

\subsection{MARL Simulation Platforms}
Effective evaluation and development of Multi-Agent Reinforcement Learning (MARL) algorithms require simulation platforms that can model complex, dynamic environments with multiple interacting agents. In the context of autonomous vehicle coordination and intelligent transportation systems (ITS), several simulation environments have become prominent for testing MARL algorithms. These platforms support integration with RL libraries, allow for custom scenario creation, and provide real-time traffic dynamics.

\begin{enumerate}
    \item \textbf{SUMO} \footnote{https://eclipse.dev/sumo/}: An open-source, microscopic traffic simulator that supports large-scale transportation networks. It is widely used for tasks such as intersection control, lane merging, and vehicle routing. MARL agents can interface with SUMO via the TraCI API.
    \\
    \item \textbf{CARLA} \footnote{https://carla.org/}: A high-fidelity 3D simulator for autonomous driving research. It provides detailed vehicle dynamics, sensor models (e.g., LIDAR, cameras), and supports integration with MARL for decision-making in urban environments such as intersections and lane changes.
    \\
    \item \textbf{CityFlow} \footnote{https://cityflow.readthedocs.io/en/latest/introduction.html}: A high-performance traffic simulator tailored for large-scale signal control environments. It is particularly suited for graph-based MARL research on coordinated intersection management.
    \\
    \item \textbf{SMARTS} \footnote{https://github.com/huawei-noah/SMARTS}: A recent platform focused on realistic multi-agent AV interactions. SMARTS offers modular scenario design and supports MARL algorithms like MAPPO and QMIX in complex environments.
    \\
    \item \textbf{Highway-env} \footnote{https://github.com/Farama-Foundation/HighwayEnv}: A lightweight simulator for highway scenarios, including lane keeping, merging, and platooning. It is widely used for prototyping and benchmarking MARL methods in constrained driving environments.
    \\
    \item \textbf{AIMSUN, VISSIM, and Paramics} \footnote{https://www.aimsun.com/}: Commercial-grade traffic simulators that support high-fidelity, city-scale modeling. These platforms are used for validating MARL-based strategies in more realistic traffic networks and are often applied in industry or urban policy research.
    \\
    \item \textbf{PRESCAN} \footnote{https://plm.sw.siemens.com/en-US/simcenter/autonomous-vehicle-solutions/prescan/}: A high-fidelity simulation platform designed for autonomous driving research, featuring photorealistic sensor modeling, traffic scenarios, and advanced vehicle dynamics. PRESCAN is widely used in industry and academia for testing MARL-based coordination strategies in safety-critical driving tasks such as intersection negotiation, obstacle avoidance, and multi-agent highway maneuvers.
    \\
    \item \textbf{MATLAB/Simulink} \footnote{https://www.mathworks.com/products/simulink.html}: A widely used engineering simulation environment offering powerful tools for modeling vehicle dynamics, control systems, and signal processing. It supports integration with Stateflow and Reinforcement Learning Toolbox, enabling the implementation and testing of MARL algorithms for tasks such as adaptive cruise control, platooning, and cooperative lane changing in a highly customizable and modular framework.
\end{enumerate}

\begin{figure*}
\centering
\subfloat[]{\includegraphics[width=2.5in]{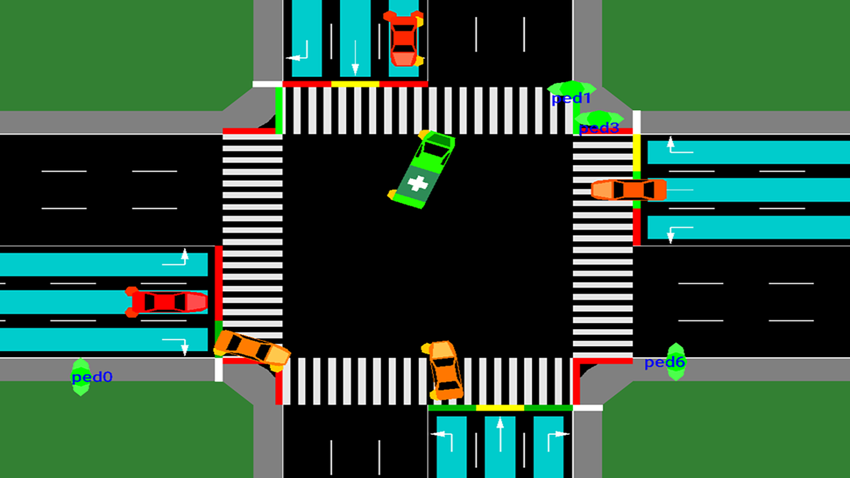}%
\label{SUMO}}
\hfil
\subfloat[]{\includegraphics[width=2.5in]{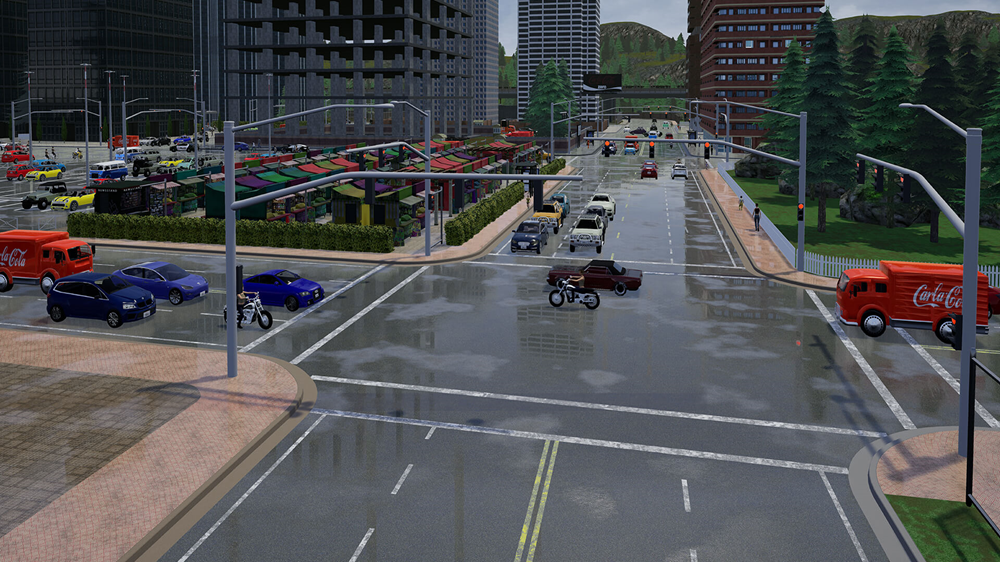}%
\label{CARLA}}
\hfil
\subfloat[]{\includegraphics[width=2.5in]{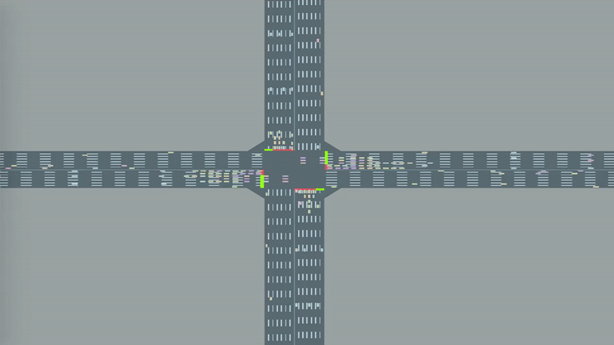}%
\label{CITYFLOW}}
\hfil
\subfloat[]{\includegraphics[width=2.5in]{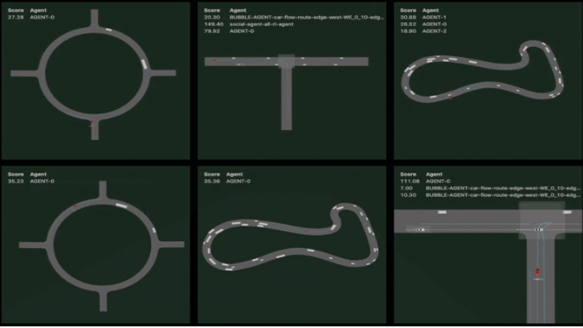}%
\label{SMARTS}}
\hfil
\subfloat[]{\includegraphics[width=2.5in]{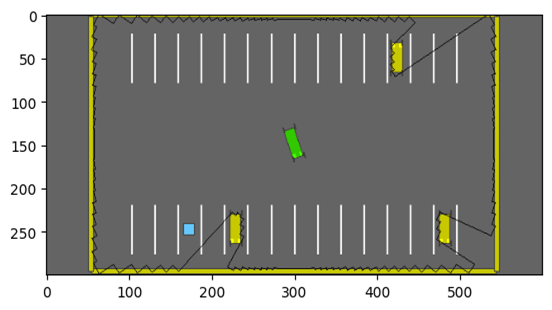}%
\label{HIGHWAY ENV}}
\hfil
\subfloat[]{\includegraphics[width=2.5in]{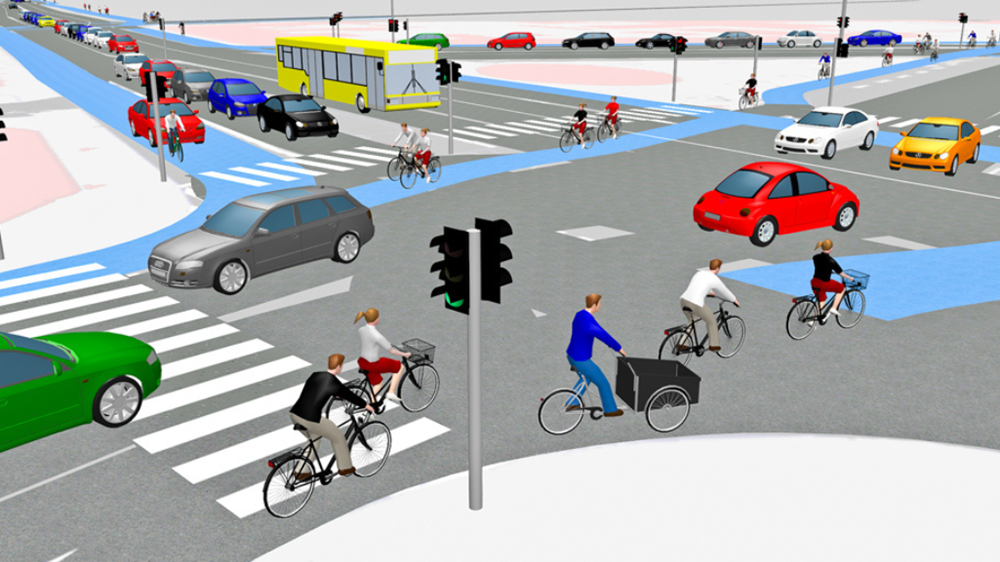}%
\label{VISSIM}}
\hfil
\subfloat[]{\includegraphics[width=2.5in]{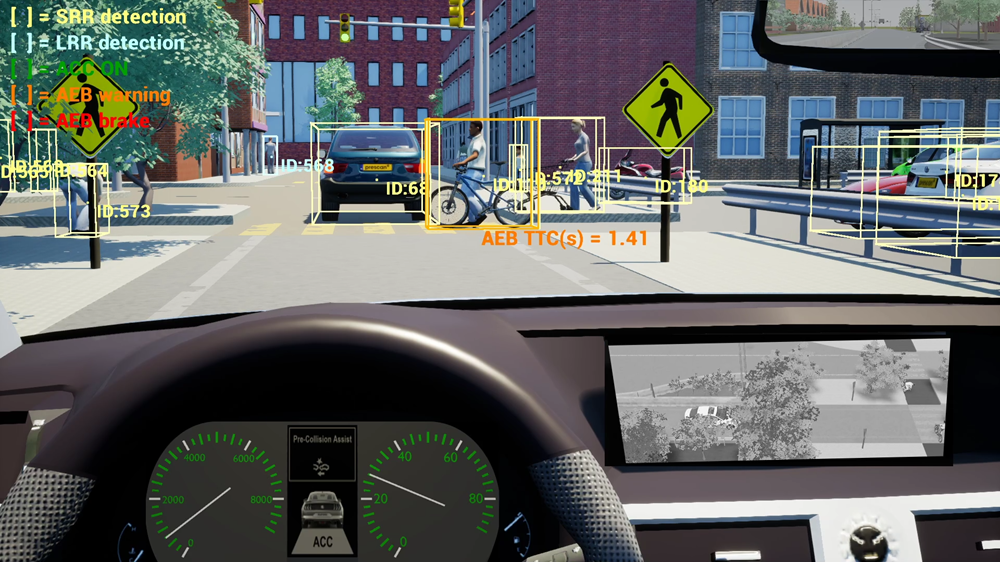}%
\label{PRESCAN}}
\hfil
\subfloat[]{\includegraphics[width=2.5in]{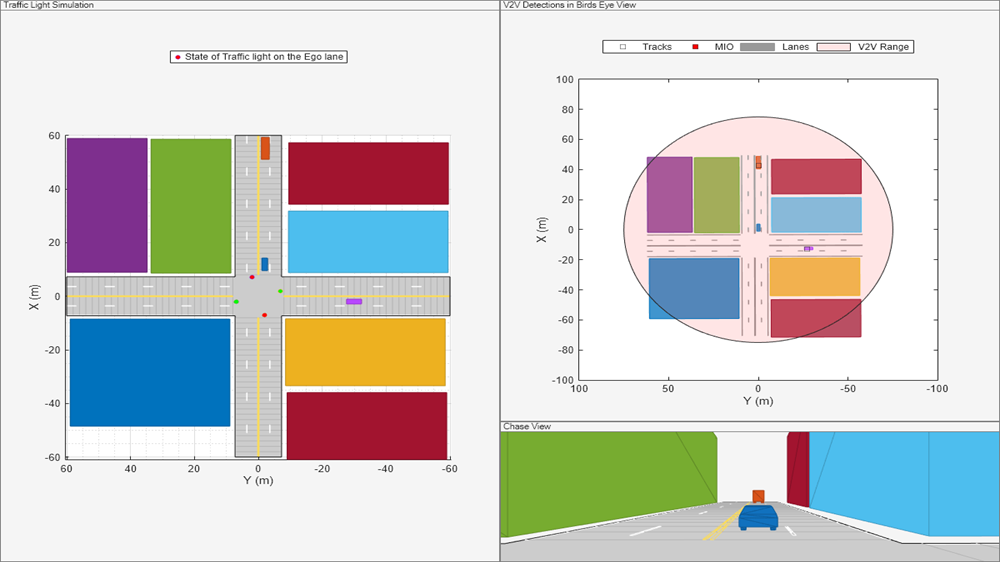}%
\label{MATLAB}}
\caption{Common Simulators (a) SUMO  (b) CARLA (c) CITYFLOW (d) SMARTS (e) HIGHWAY ENV (f) VISSIM (g) PRESCAN (h) MATLAB}
\label{fig_sim}
\end{figure*}

These simulation platforms shown in Figure \ref{fig_sim} form the backbone of experimental validation in MARL research. Selecting an appropriate environment depends on the specific task (e.g., lane merging vs. platooning), scale (e.g., intersection vs. city-wide), and required fidelity (e.g., sensor-level vs. abstracted traffic dynamics).

\section{APPLICATION OF MARL IN INTELLIGENT TRANSPORTATION SYSTEMS} \label{applications}
Multi-Agent Reinforcement Learning (MARL) has emerged as a powerful paradigm for solving complex, dynamic problems in intelligent transportation systems (ITS), where multiple decision-makers vehicles, signals, fleets, or aerial agents must coordinate under uncertainty \cite{huang2024multi}. MARL’s ability to handle distributed decision-making, adapt to real-time data, and scale to large environments makes it especially suitable for managing the growing complexity of modern mobility networks \cite{qin2021multi}. This section explores its key applications across various ITS domains.

\subsection{Traffic Signal Control}
One of the earliest and most studied applications of MARL in ITS is traffic signal control, where intersections are modeled as agents that learn to minimize congestion and delay \cite{kolat2023multi}. In single intersection control, MARL agents learn optimal phase timings based on local traffic states, such as queue lengths or vehicle waiting times \cite{kolat2023multi}. While effective for localized improvements, this approach is limited in its capacity to optimize flows at a network level \cite{yang2023causal}.

To address this, network-wide control approaches model each intersection as a cooperative agent, where coordination is key \cite{kolat2023multi}. Techniques such as Graph-based MARL (e.g., CoLight, PressLight) allow intersections to exchange information and adjust their strategies based on downstream and upstream traffic dynamics \cite{yang2023causal}. Coordination strategies like hierarchical learning, value decomposition, or message-passing enable emergent traffic patterns such as green waves and adaptive lane prioritization \cite{yang2023causal}.

These methods support real-time control by adapting to fluctuating traffic conditions. MARL-based systems can reduce average delays, improve throughput, and dynamically respond to incidents or surges, outperforming traditional rule-based or fixed-timing approaches. A summary of recent MARL papers in TSC is given in Table \ref{tab:marl_tsc}

\begin{sidewaystable*}
\centering
\caption{Summary of Multi-Agent RL Approaches for Traffic Signal Control (TSC)} \label{tab:marl_tsc}
\begin{tabular}{p{0.5cm}p{4.5cm}p{4cm}p{4cm}p{4.5cm}p{3.5cm}}
\hline
\textbf{Paper} & \textbf{Main Contribution} & \textbf{MARL Algorithm} & \textbf{Objective} & \textbf{Outcome} & \textbf{Simulator}  \\ \hline

\cite{wang2021adaptive} & Introduced a scalable multi-agent framework for adaptive traffic signal control using group-based coordination. & Cooperative Group-Based Multi-Agent Q-Learning (CGB-MAQL) & Improve coordination and scalability in large-scale TSC & Better congestion mitigation and energy efficiency & SUMO \\
\\

\cite{chu2019multi} & Decentralized A2C with spatial discounting and neighbor awareness  & Multi-agent A2C (MA2C) & Stabilize learning and improve policy in large scale TSC & Outperformed Independent A2C (IA2C) and independent Q-learning (IQL) in control efficiency & SUMO \\
\\

\cite{prabuchandran2014multi} & Multi-agent Q-learning with simple cost-based coordination  & Q-learning  & Reduce average vehicle delay at intersections & Outperformed fixed and adaptive baselines & VISSIM (real road networks) \\
\\

\cite{van2016coordinated} & DQN with coordination via max-plus; focus on reward shaping and stability & DQN with coordination & Improve multi-agent policy coordination in TSC & Coordination improved travel time; noted instability in single-agent cases & SUMO \\
\\

\cite{chandra2022gameplan} & Introduced GAMEPLAN, a game-theoretic auction system prioritizing agents based on observable driving behavior to ensure collision-free planning. & Game-theoretic MARL & Turn-based ordering for unsignalized traffic scenarios & 10–20\% fewer collisions than DRL; 3.5\% better than state-of-the-art auctions & Real-world + synthetic (merging/intersections/roundabouts) \\
\\

\cite{cui2021scalable} & Developed modular and distributed MADRL policies for large-scale traffic networks with congestion reduction. & Distributed MADRL + Transfer RL & Maximize vehicle outflow and reduce congestion & Improved over human traffic; scalable to hundreds of AVs & SUMO + RLlib \\
\\

\cite{chen2020delay}  & Proposed Delay-Aware Markov Games with centralized training, decentralized execution to mitigate delay effects. & Delay-Aware MARL with CTDE & Improve stability and performance under delay in cooperative/competitive settings & Outperformed standard MARL in delayed environments & Multi-Agent Particle Environment\\
\\
\cite{xu2021leveraging} & Proposed MA-DRLS for cooperative control at nonsignalized intersections using FCTP scheduling and DRL. & MA-DRLS (Multi-agent DRL) & Throughput maximization and wait time reduction & Significantly better throughput and lower wait time than traffic light methods & Custom intersection simulation with V2X \\
\\
\cite{antonio2022multi} & Presented adv.RAIM using end-to-end MADRL with curriculum self-play and LSTM-based control. & adv.RAIM (End-to-End MADRL) & Eliminate collisions, reduce waiting and travel time at intersections & Reduced travel time 59\%, congestion delay 95\%, waiting time 88\% & Custom AIM simulator (3-lane, 1200 veh/h/lane) \\
\hline
\end{tabular}
\end{sidewaystable*}

\subsection{Autonomous Vehicle Coordination}
In the context of connected and autonomous vehicles (CAVs), MARL provides a decentralized framework for vehicle coordination \cite{hua2025multi}. Applications include lane merging on highways, intersection crossing without traffic lights, and roundabout negotiation, where each vehicle acts as an agent that must infer the intentions of others and optimize for safety and efficiency \cite{nakka2022multi}.

In platooning and convoy control, groups of autonomous vehicles travel closely together to improve aerodynamics and traffic flow \cite{yadav2023comprehensive}. MARL enables these vehicles to jointly learn policies that minimize fuel consumption while maintaining safety and communication constraints \cite{vinitsky2023optimizing}. Adaptive gap control and coordinated acceleration/deceleration patterns are examples of emergent behaviors learned through MARL \cite{han2023multi}.

Multi-agent highway driving simulations, such as those used in CARLA, SUMO, or Flow, help train and validate MARL policies in realistic traffic scenarios \cite{yadav2023comprehensive, okafor2021electric}. These simulations test cooperative and competitive interactions between human-driven and autonomous vehicles, ensuring safe deployment of MARL in real-world autonomous traffic \cite{vinitsky2023optimizing}. A summary of recent MARL papers in autonomous vehicle coordination and control is given in Table \ref{Tab:marl_avc}

\begin{sidewaystable*}
\centering
\footnotesize
\caption{Summary of MARL-Based Publications for Autonomous Vehicle Coordination} \label{Tab:marl_avc}
\begin{tabular}{p{0.5cm}p{4.5cm}p{4cm}p{4cm}p{4.5cm}p{3.8cm}}
\hline
\textbf{Paper} & \textbf{Main Contribution} & \textbf{MARL Algorithm} & \textbf{Objective} & \textbf{Outcome} & \textbf{Simulator} \\
\hline
\cite{gong2024cooperative} & Introduced a method for cooperative planning at roundabouts & Adaptive Monte Carlo Tree Search (AMCTS) & Time-optimal, collision-free trajectory planning & Efficient and safe roundabout traversal & Custom Simulator \\
\\
\cite{xu2024multi}  & Proposes Relative Position Encoding - Multi-Actor Attention Critic (RPE-MAAC) for stabilizing mixed platoons with CAVs and HDVs & RPE-MAAC & Mixed platoon stability and safety & Enhanced comfort and reduced disruptions & Custom numerical simulator \\
\\
\cite{peake2020multi} & Proposes an cooperative adaptive cruise control (CACC) MARL framework using LSTM with learned communication protocol & Policy Gradient with LSTM for MARL-CACC) & String stability and communication robustness & Improved trajectory and convergence & Custom Simulator \\
\\
\cite{acquarone2023cooperative} & Applies TD3 for energy-efficient CACC in heavy-duty BEVs & TD3  & Energy savings with safety and comfort & Up to 19.8\% energy reduction with comfort preserved & HHDDT driving cycle simulation \\
\\
\\
\cite{zhou2022multi} & Presents Multi-agent Advantage Actor-Critic (MA2C) for cooperative lane-changing in mixed AV/HDV traffic & MA2C & Safe, comfortable, fuel-efficient lane-changing & Superior to benchmarks in mixed traffic & Custom highway simulation environment \\
\\
\cite{leurent2020safe} & Explores robust and efficient behavioral planning with risk-sensitive RL & Budgeted RL, Tree-based Planning & Balancing safety and efficiency & Continuum of behaviors from conservative to aggressive & Highway-env (open-source environment) \\
\\

\cite{chen2021graph} & Graph convolutional Q-network for multi-agent CAV cooperative control & Graph Q-Network (GQN) & Efficient lane-changing in high-density traffic & Improved safety and mobility compared to rule-based and LSTM fusion methods & Custom simulator \\
\\
\cite{chen2017decentralized} & Decentralized collision avoidance without communication via value network & Value network & Collision-free multi-agent path planning under partial observability & 26\% improvement over ORCA in navigation efficiency & 2D simulation environment \\
\\
\cite{shalev2016safe} & Safe hierarchical RL via decomposition of Desires and trajectory planning & Policy Gradient  & Learning driving policy with guarantees on safety and comfort & Low-variance RL strategy for autonomous driving under uncertainty & Not specified \\
\\
\cite{troullinos2021collaborative} & Lane-free CAV coordination using coordination graphs and max-plus & Max-plus & Coordination without lanes to increase traffic flow and safety & Higher speeds and flow rates with efficient lateral usage & SUMO  \\
\\
\cite{thakkar2024hierarchical} & Hierarchical control for head-to-head autonomous racing & Hierarchical RL with planning and control layers & Competitive racing behavior with safety & More robust and aggressive racing strategies & F1TENTH simulator \\
\\
\cite{toghi2021altruistic}  & Inverse RL-based multi-agent lane merging behaviors & IRL with deep policies & Imitate human-like merging decisions in CAVs & Learned policies outperform rule-based merging & CARLA \\
\\
\cite{li2021reinforcement} & Energy-aware platoon control using DRL & Actor-Critic  & Reduce energy loss during traffic oscillations & Smoother velocity profiles and less energy use & Custom simulator \\
\hline

\end{tabular}
\label{tab:MARL_summary}
\end{sidewaystable*}

Multi-agent reinforcement learning has also been employed in various domains within intelligent transportation systems, including freight and logistics, UAV control and coordination, and transportation safety, among others. A summary of recent studies in these areas is presented in Table \ref{Tab:marl_its_sum}.


\begin{sidewaystable*}
\centering
\caption{Key Papers on MARL for ITS} \label{Tab:marl_its_sum}
\begin{tabular}{p{0.5cm}p{2.5cm}p{6cm}p{4cm}p{4cm}p{4cm}}
\hline
\textbf{Paper} & \textbf{Category} & \textbf{Main Contribution} & \textbf{MARL Algorithm} & \textbf{Problem Domain} & \textbf{Simulation Environment} \\
\hline
\cite{cui2019multi}  &  & Resource allocation for multi-UAV downlink networks without information exchange between agents. & Independent Q-Learning & Power control, user \& subchannel selection & Custom simulator \\
\\
\cite{qie2019joint} &  & Joint target assignment and path planning using MADDPG. & MADDPG  & Target allocation \& path planning & Custom 2D dynamic simulator \\
\\
\cite{pham2018cooperative}  & UAV & Field coverage by a UAV team while minimizing overlap using correlated equilibrium. & Game-theoretic MARL with function approximation & Coverage optimization & Physical and simulated testbeds \\
\\
\cite{shamsoshoara2019distributed} &  & Cooperative spectrum sharing with task allocation in constrained communication settings. & Decentralized Q-Learning & Task division: relaying vs. sensing & MATLAB-based simulator \\
\\
\cite{jung2021coordinated}  &  & Energy-aware UAV charging via coordinated deep MARL using CommNet. & CommNet-based MADRL & Charging resource allocation & Simulated smart grid environment \\
\\
\hline
\\
\cite{lin2018efficient} &  & Introduced contextual MARL with scalable coordination for fleet management. & CA2C, CDQN & Ride-hailing fleet repositioning & Custom imulator with Didi Chuxing data \\
\\
\cite{shalev2016safe} & Automotive & Decomposed policy into safe planning and learnable desires using Option Graph. & Non-Markovian Policy Gradient & Autonomous driving strategy & Custom double-merge scenario \\
\\
\hline
\\

\cite{li2019cooperative} &  & Proposes cooperative MARL for resource balancing in logistics with cooperative reward shaping. & Custom Cooperative MARL & Ocean container repositioning & Simulated ocean freight network \\
\\
\cite{saifullah2024multi} &  & Uses DDMAC-CTDE for lifecycle management of transport infrastructure. & DDMAC-CTDE & Transportation infrastructure I\&M & Custom simulator \\
\\
\cite{van2022strategic} & Freight and Logistics & MARL for decentralized bidding in freight transport markets. & Policy Gradient  & Freight bidding strategy & Custom simulator \\
\\
\cite{wang2025multi} &  & Shared MARL for dynamic logistics service collaboration. & Multi-Agent A2C  & Freight collaboration & Custom simulator \\
\\
\hline
\end{tabular}
\label{tab:marl_uav_summaryy}
\end{sidewaystable*}

\begin{sidewaystable*}
\centering
\begin{tabular}{p{0.5cm}p{2.5cm}p{6cm}p{4cm}p{4cm}p{4cm}}
\hline
\textbf{Paper} & \textbf{Category} & \textbf{Main Contribution} & \textbf{MARL Algorithm} & \textbf{Problem Domain} & \textbf{Simulation Environment} \\

\hline
\\
\cite{elsayed2021safe} &  & Introduced MADRL framework that effectively addresses dynamic flexible assembly job shop scheduling (FAJSS) problems under uncertainty in processing and transport times &  MADDPG & Flexible assembly job shop scheduling & Custom Discrete-Event Simulators \\
\\
\cite{bernhard2020bark} & Safety & Proposes BARK, a benchmark for safety-oriented evaluation of MARL policies. & Independent PPO, MADDPG, QMix & Safety benchmarking across environments & MiniGrid, Safety-Gymnasium \\
\\
\cite{zhu2022can} &  & Hybrid RL model for AV motion planning at unsignalized mid-block crosswalks. & Policy-Gradient RL (hybrid model) & Pedestrian-aware AV driving & Real-world pedestrian speed profiles \\
\\
\cite{bautista2022autonomous} &  & RL-MPC integration for safe intersection crossing & TD3 & Urban intersections & Semantic map-based simulator \\

\hline
\\
\cite{jia2022multi} & & Joint trajectory prediction using egocentric and allocentric views to enable symmetrical multi-agent modeling with GNN & MADDPG & Multi-agent trajectory prediction (vehicles \& pedestrians) & Custom simulator \\
\\
\cite{ma2021continual} &  & Introduces a conditional generative memory for continual multi-agent prediction avoiding catastrophic forgetting & MADDPG & Continual multi-agent interaction behavior prediction & SUMO \\
\\
\cite{xie2021learning} & Trajectory Prediction & Latent strategy learning and influence modeling for co-adaptive multi-agent interaction & QMIX & Non-stationary multi-agent interaction & Custom simulator \\
\\
\cite{wiederer2022anomaly} &  & Survey and experimental framework for modeling cooperation in mixed human-machine environments & Multiple RL architectures (reviewed) & Cooperative multi-agent learning & Custom simulator \\
\\
\cite{mo2022multi} &  & Introduces framework for cooperative control in mixed traffic with attention-based feature integration & Policy gradient & Mixed traffic cooperative control & SUMO-based traffic simulation \\
\hline
\end{tabular}
\label{tab:marl_uav_summary}
\end{sidewaystable*}

\section{CHALLENGES IN MARL FOR ITS} \label{challenges}
Multi-Agent Reinforcement Learning (MARL) offers substantial promise for improving the performance and adaptability of Intelligent Transportation Systems (ITS). From managing traffic flows and coordinating fleets to optimizing logistics networks and autonomous vehicle behavior, MARL can enable intelligent decision-making across distributed, dynamic environments. However, transitioning these capabilities from theory to real-world application presents a host of complex and interrelated challenges.

A fundamental challenge lies in the issue of scalability. ITS environments often consist of a large number of interacting agents such as vehicles, traffic signals, or delivery drones operating concurrently. As the number of agents increases, the joint state-action space grows exponentially, making centralized control or joint-policy learning computationally infeasible. This combinatorial explosion necessitates the use of decentralized learning architectures or factorized representations to maintain tractability in large-scale scenarios like network-wide traffic signal coordination or city-level mobility management.

Another major hurdle is credit assignment in cooperative MARL settings. When agents collectively contribute to a global objective, such as minimizing congestion or maximizing throughput, it becomes difficult to determine the contribution of each individual agent to the overall outcome. Inaccurate credit assignment can lead to inefficient or misguided learning, particularly in systems with heterogeneous agents that play different roles. Approaches like value function factorization (e.g., QMIX) and counterfactual baselines (e.g., COMA) attempt to tackle this issue by better estimating individual agent contributions.

Many ITS tasks also involve continuous control variables, such as vehicle acceleration, steering angles, or lane-changing maneuvers. Learning effective policies in continuous action spaces is significantly more challenging than in discrete settings. Algorithms like MADDPG and MAPPO have been developed to address this, but they often require careful parameter tuning, are sensitive to stochastic noise, and may suffer from instability especially when scaled to high-dimensional, multi-agent contexts.

Communication between agents is another vital yet difficult aspect of MARL in ITS. Effective coordination frequently relies on the timely exchange of information between agents, such as vehicle-to-vehicle or vehicle-to-infrastructure messages. However, real-world communication is constrained by limited bandwidth, transmission delays, packet loss, and unreliable connectivity. Designing learning-based communication protocols that are robust, efficient, and scalable remains an active area of research, especially for applications like swarm coordination, cooperative merging, and platooning.

Beyond these algorithmic complexities, the learning process itself presents formidable challenges. ITS scenarios typically involve high-dimensional sensor inputs, rapidly changing environments, and multiple competing goals. Developing MARL agents that can learn effectively in such conditions requires substantial computational resources, meticulous reward design, and extensive tuning of learning parameters. Moreover, the policies must not only succeed in training but also generalize across diverse real-world situations, such as varying traffic densities, unexpected detours, or weather-related disruptions.

Lastly, real-world deployment of MARL in transportation systems must address non-technical constraints such as safety assurance, policy explainability, and operational robustness. A significant gap often exists between performance in simulation and deployment in physical systems, known as the "sim-to-real" gap. This arises due to discrepancies in sensing accuracy, environment dynamics, and agent behaviors. Bridging this gap requires methods like domain randomization, real-world fine-tuning, or adaptive online learning to ensure that trained policies remain effective and safe under real-world conditions.

Collectively, these challenges highlight the complexity of deploying MARL in ITS and underscore the need for continued interdisciplinary research that integrates advances in machine learning, systems engineering, and transportation science.

\section{FUTURE RESEARCH DIRECTIONS} \label{future}

As Multi-Agent Reinforcement Learning (MARL) becomes increasingly integrated into Intelligent Transportation Systems (ITS), numerous promising research directions have emerged to overcome current limitations and unlock greater potential. These opportunities span theoretical frameworks, algorithmic innovations, and practical deployments, reflecting the complex and evolving nature of ITS environments.

One pressing area for future research is the development of safe and explainable MARL systems. Safety is a critical concern in transportation, and there is an urgent need for MARL algorithms that can offer formal guarantees under uncertainty while adhering to safety constraints. Incorporating methods from safe reinforcement learning, constrained Markov Decision Processes (MDPs), and formal shielding techniques can enhance the reliability of MARL in high-stakes environments. At the same time, explainability is essential for trust and adoption. Mechanisms such as interpretable policy models, causal reasoning, and human-in-the-loop learning can improve transparency and help stakeholders better understand and validate agent decisions.

Another key challenge lies in sim-to-real transfer and domain adaptation. Policies trained in simulated environments often degrade in performance when deployed in the real world due to discrepancies in dynamics, noise, or context known as the "sim-to-real" gap. Bridging this divide requires techniques like domain randomization, curriculum learning, and adaptive fine-tuning. Moreover, multi-fidelity simulations that integrate both high- and low-resolution models can help expose agents to a broader range of scenarios, improving generalization and robustness in real-world applications.

Future research must also address the need for multi-objective and human-centric learning approaches. Transportation systems are inherently multi-faceted, requiring agents to manage trade-offs among efficiency, equity, environmental sustainability, and passenger comfort. Developing MARL frameworks capable of optimizing across multiple, often conflicting objectives will be essential. Additionally, integrating models of human decision-making such as bounded rationality, social norms, and user preferences can produce more realistic and socially aligned agent behaviors that better reflect how people interact with transportation systems.

As ITS increasingly functions in distributed settings, communication-efficient and decentralized MARL methods are becoming more critical. In many scenarios, agents operate in bandwidth-constrained or intermittent communication environments. Research must focus on enabling agents to learn what, when, and with whom to communicate effectively. Innovations such as emergent communication protocols, attention-based messaging systems, and decentralized policy architectures that leverage latent state representations offer promising paths forward.

Lastly, generalization and lifelong learning remain foundational challenges for MARL in ITS. Given the dynamic nature of transportation networks, agents must continuously adapt to new cities, evolving infrastructure, changing traffic patterns, and unforeseen events without retraining from scratch. Strategies such as continual learning, meta-learning, few-shot adaptation, and transfer-based pretraining can equip agents with the flexibility to respond to novel tasks and maintain long-term performance across diverse and shifting environments.

Together, these research directions aim to make MARL a more powerful, reliable, and practical tool for next-generation transportation systems capable of delivering safe, adaptive, and intelligent decision-making at scale.

\section{CONCLUSION} \label{conclusion}
Multi-Agent Reinforcement Learning (MARL) has emerged as a powerful tool for enhancing the adaptability, scalability, and autonomy of Intelligent Transportation Systems (ITS). Through coordinated decision-making and learning in decentralized environments, MARL enables various transportation agents such as vehicles, traffic lights, and delivery systems to collaboratively address the complex and dynamic challenges of modern mobility networks.

This paper has provided a comprehensive overview of recent developments in applying MARL to ITS, highlighting its potential across multiple application domains including traffic management, freight logistics, UAV coordination, and autonomous vehicle interactions. In doing so, it has also identified the fundamental challenges that hinder real-world deployment, such as scalability issues, credit assignment complexity, communication constraints, and the sim-to-real gap.

To bridge these gaps, future research must focus on designing safe, explainable, and human-centric MARL frameworks that can generalize across environments and support real-time operation under uncertainty. Emphasis on communication efficiency, domain adaptation, and lifelong learning will further ensure the robustness and practicality of MARL in large-scale ITS deployments. By addressing these challenges through interdisciplinary collaboration and advanced algorithmic innovations, MARL can play a pivotal role in shaping the future of intelligent, resilient, and sustainable transportation systems.



\printcredits

\bibliographystyle{cas-model2-names}

\bibliography{cas-refs}





\end{document}